\documentclass[journal]{IEEEtran}

\usepackage[ruled,lined,commentsnumbered]{algorithm2e} 
\usepackage{graphicx} 
\usepackage{multirow}
\usepackage{adjustbox}
\usepackage{amsmath}
\usepackage{url}
\usepackage{booktabs}

\usepackage{epsfig}
\usepackage{graphicx}
\usepackage{amsmath}
\usepackage{amsfonts}
\usepackage{amssymb}
\usepackage{array}
\usepackage{multirow}
\usepackage{rotating}
\usepackage{epstopdf}
\usepackage{caption}
\usepackage{subcaption}
\usepackage[dvipsnames]{xcolor}
\usepackage{adjustbox}
\usepackage{titling}

\usepackage{bbding}
\usepackage{pifont}
\usepackage{wasysym}
\usepackage{amssymb}

\usepackage{afterpage}

\begin{document}
%
\title{Fingerprint Presentation Attack Detection Based on Local Features Encoding for Unknown Attacks}
\date{}%
%
%

\author{
	L\'azaro J. Gonz\'alez-Soler\thanks{L. J. Gonz\'alez-Soler, M. Gomez-Barrero and C. Busch are with the da/sec - Biometrics and Internet Security Research Group, Hochschule Darmstadt, Germany (e-mail: \{lazaro-janier.gonzalez-soler,marta.gomez-barrero,christoph.busch\}@h-da.de).
	L. Chang is with the Tecnologico de Monterrey, School of Engineering and Science, Mexico (e-mail: lchang@tec.mx).
	A. P\'{e}rez-Su\'{a}rez is with the Advanced Technologies Application Center (CENATAV), La Habana, Cuba (e-mail: asuarez@cenatav.co.cu).
	This work was partially supported by the German Federal Ministry of Education and Research (BMBF) as well as by the Hessen State Ministry for Higher Education, Research and the Arts (HMWK) within the Center for Research in Security and Privacy (CRISP, www.crisp-da.de)}, 
	Marta Gomez-Barrero,
	Leonardo Chang,
	Airel P\'{e}rez-Su\'{a}rez,
	Christoph Busch
}

\markboth{}%
{Gonzalez-Soler \MakeLowercase{\textit{et al.}}: Fingerprint PAD Based on Local Features Encoding for Unknown Attacks}
%



\maketitle

\begin{abstract}
Fingerprint-based biometric systems have experienced a large development in the last years. Despite their many advantages, they are still vulnerable to presentation attacks (PAs). Therefore, the task of determining whether a sample stems from a live subject (i.e., bona fide) or from an artificial replica is a mandatory issue which has received a lot of attention recently. Nowadays, when the materials for the fabrication of the Presentation Attack Instruments (PAIs) have been used to train the PA Detection (PAD) methods, the PAIs can be successfully identified. However, current PAD methods still face difficulties detecting PAIs built from unknown materials or captured using other sensors. Based on that fact, we propose a new PAD technique based on three image representation approaches combining local and global information of the fingerprint. By transforming these representations into a common feature space, we can correctly discriminate bona fide from attack presentations in the aforementioned scenarios. The experimental evaluation of our proposal over the LivDet 2011 to 2015 databases, yielded error rates outperforming the top state-of-the-art results by up to 50\% in the most challenging scenarios. In addition, the best configuration achieved the best results in the LivDet 2019 competition (overall accuracy of 96.17\%). 
\end{abstract}


%

\section{Introduction}   
\label{sec:introduction}

Biometric recognition is based on the use of distinctive anatomical and behavioural characteristics to automatically recognise a subject \cite{maltoni2009handbook}. Among other biometric characteristics, fingerprints offer a high recognition accuracy and at the same time enjoy a high popular acceptance. Despite these and other advantages, fingerprint-based recognition systems can be circumvented by launching Presentation Attacks (PAs), in which an artificial fingerprint, denoted as Presentation Attack Instrument (PAI) is presented to a sensor \cite{galbally2006vulnerability,galbally2010evaluation,matsumoto2002impact,ISO-IEC-30107-3-PAD-metrics-170227}. 



The threat posed by PAIs is not reduced to an academic issue. In 2002, Matsumoto \textit{et al.} \cite{matsumoto2002impact,matsumoto2002gummy} analysed the vulnerabilities of eleven commercial fingerprint-based biometric systems to gummy fingerprints. The experimental evaluation showed that $68\%$ to $100\%$ of the PAIs built with cooperative methods were accepted as bona fide presentations (i.e., genuine or live fingers). In 2009, Japan reported the use of presentation attacks in one of its airports, and in 2013, a Brazilian doctor used artificial silicone fingerprints to tamper a biometric attendance system at the Sao Paulo hospital \cite{schuckers2016presentations}. 

In order to tackle those severe security issues, the development of Presentation Attack Detection (PAD) techniques, which automatically detect PAIs presented to the biometric capture device, is a mandatory task, which has attracted a lot of attention within the biometric research community not only for fingerprint systems \cite{marasco2015surveyAntiSpoofFp,sousedik2014fingerprintPAD}, but also for other characteristics such as face \cite{galbally2014facePADsurvey} or iris \cite{galbally16spoofingIrisChapter}. These PAD methods can be widely classified as hardware- or software-based approaches. Whereas the former require dedicated, and mostly expensive, specific hardware, software-based approaches focus on dynamic or static characteristics extracted from the same biometric samples used for recognition purposes. Therefore, software-based methods are less expensive, and will be the focus of this article.


The newest fingerprint PAD techniques based on deep learning and textural features have shown to be a powerful tool to detect most PAIs \cite{chugh2017fingerprint,chugh2018fingerprint,nogueira2016fingerprint,pala2017deep}. However, they share a common limitation: they depend both on $i)$ the material used for fabricating the PAIs, and $ii)$ the sensor used for acquiring the fingerprint samples. More specifically, their error rates are multiplied five to 18 times when either the PAIs' materials or the sensors utilised are not known a priori (see Table~\ref{tab:SOTA_ACE}). 


\begin{table}[t!]
	\centering
	\caption{Summary of the best state-of-the-art results on different scenarios. }
	\label{tab:SOTA_ACE}
	\scriptsize
	\begin{tabular}{l c c c } 
		\toprule
		\textbf{CNN} & \textbf{Known-env} & \textbf{Cross-sensor} & \textbf{Cross-DB} \\		
		\midrule
MobileNet-v1~\cite{chugh2018fingerprint}          &   0.97\%    &  14.59\%   &   17.91\%             \\
Inception-v3~\cite{chugh2017fingerprint}       	  &   1.39\%    &  16.60\%   &   18.90\%       	     \\
Deep Triple Embedding~\cite{pala2017deep}		  &   3.33\%    &  25.25\%   &   15.20\%       	     \\
VGG~\cite{nogueira2016fingerprint}			 	  &	  4.61\%    &  19.80\%	 &	 30.70\%    		 \\
		\bottomrule
	\end{tabular}\vspace*{-0.3cm}
\end{table}

To address the issue of generalisation to unknown factors, we analyse the combination of local features (i.e., Scale-Invariant Feature Transform, SIFT \cite{lowe2004distinctive}) with three different general purpose feature encoding approaches, which have shown remarkable results in object classification tasks \cite{Csurka04,peng2016bag,lampert2014attribute}: $i)$ Bag of Words (BoW), $ii)$ Vector of Locally Aggregated Descriptors (Vlad), and $iii)$ Fisher Vector (FV). The local descriptors, computed over the image gradient, allow capturing different artefacts produced by materials used for building the PAIs. Then, the afforementioned encoding approaches assign each local descriptor (i.e., SIFT) to the closest entry in a \textit{visual vocabulary} \cite{sanchez2013image}. This visual vocabulary defines a common feature space, thereby allowing a better generalisation to unknown attacks or capture devices. 

In order to evaluate the performance of the proposed methods and to allow the reproducibility of the results, we conduct a thorough experimental evaluation on the LivDet 2011, LivDet 2013, and LivDet 2015 databases. The performance is reported in compliance with the ISO/IEC 30107 international standard on PAD evaluation \cite{ISO-IEC-30107-3-PAD-metrics-170227}, thereby allowing a rigorous analysis of the results. The evaluation shows the capacity of the new method to be used in high security applications: for a high security operating point with an Attack Presentation Classification Error Rate (APCER) of 1\%, an average Bona Fide Presentation Classification Error Rate (BPCER) of 0.25\%, 0.38\% and 7.11\% was achieved, respectively, on the three databases, thereby outperforming the state-of-the-art. In addition, we would like to highlight that the proposed method took part in the Fingerprint Liveness Detection Competition 2019, achieving the best detection performance with an average accuracy of 96.17\%~\cite{Orru-LivDet2019-arxiv-2019}.

The remainder of this paper is organized as follows: related works are summarised in Sect.~\ref{sec:related_work}. In Sect.~\ref{sec:proposals}, we describe the proposed PAD methods. The experimental evaluation is presented in Sect.~\ref{sec:Results}. Finally, conclusions and future work directions are presented in Sect.~\ref{sec:conclusions}.     

\section{Related work}   
\label{sec:related_work}

\begin{figure*}[t!]
\centering
\includegraphics[width = \linewidth]{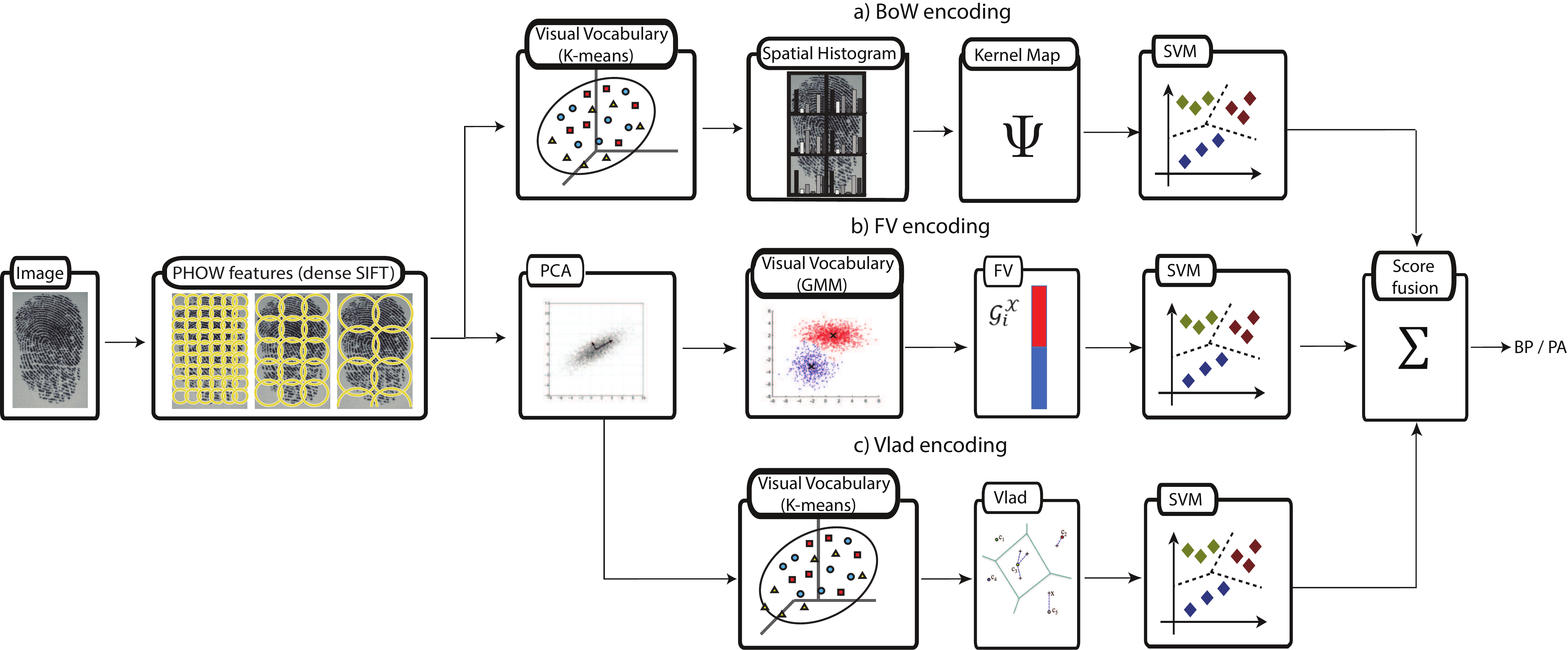}
\caption{PAD approach overview. First, dense-SIFT descriptors are computed at different scales. Then this features are encoded using a previously learned visual vocabulary by means of three different approaches: a) BoW, b) FV, and c) Vlad. Afterwards, the fingerprint descriptor is classified using a linear SVM. The final decision is made by a weighted sum rule-based fusion.}\vspace*{-0.5cm}
\label{fig:General_overview}
\end{figure*}

As we mentioned in Sect.~\ref{sec:introduction}, we focus on static software-based fingerprint PAD methods, since they are the most time and cost efficient. In particular, we review those methods based on either deep learning or addressing scenarios with unknown factors. For more details on other methods, the reader is referred to \cite{marasco2015surveyAntiSpoofFp,sousedik2014fingerprintPAD,Tolosana-FingerprintSWIR-CNN-PAD-arxiv-2019}.

In this context, it has been observed that some textural properties including the morphology, smoothness, and ridge-valley structure may be different between attack and bona fide presentations, and can thus be used to discriminate them. Building upon this idea, several texture-based PAD methods have been proposed in the literature \cite{jia2014multi,gragnaniello2015local}. More recently, new methods based on deep learning approaches have significantly outperformed any earlier PAD techniques. For instance, Nogueira \textit{et al.} \cite{nogueira2016fingerprint} benchmarked three classic Convolutional Neural Networks (CNN). One of their proposals achieved the best results in the LivDet 2015 competition, with an overall accuracy of $95.5\%$. In spite of those promising results, the main limitation of these methods is that they learn features from a whole image with a fixed size. In many cases, also within the LivDet databases, the Region of Interest (ROI) covers only a small area of the whole image (e.g., 19\% for some subsets of LivDet 2011), thus not being large enough to allow an efficient PA detection. This is highlighted by the results achieved on the LivDet 2011 - Italdata dataset, where the ACER increased up to $9.2\%$.





To address the small ROI issue, Pala and Bhanu \cite{pala2017deep} proposed training a triple convolutional network on one fixed size and randomly extracted patch per image. In spite of the obtained improvement with respect to the previous whole-image-based approach \cite{nogueira2016fingerprint}, in the random patch extraction process several patches extracted from Italdata 2011 could stem from the background region of the image, thereby resulting in a still high ACER of 5.1\%.

More recently, and based on the fact that PAIs produce spurious minutiae on a fingerprint image, Chugh \textit{et al.} \cite{chugh2017fingerprint,chugh2018fingerprint} proposed a deep learning framework for independently classifying local patches around minutiae extracted from a fingerprint image. 
The final bona fide vs PA decision was defined as the average between PAD scores of the local patches. This approach additionally allows finding PA regions inside a sample, even if the PAI only covers part of the underlying fingerprint. The method achieves the lowest ACER values reported so far over the LivDet databases (see Table~\ref{tab:SOTA_ACE}, left column). However, despite the excellent results reported in the known environment (i.e., known attacks and known sensors), an evaluation on more challenging scenarios (i.e., unknown sensors and/or PAI fabrication materials) shows an increase in the error rates (see Table~\ref{tab:SOTA_ACE}).  

Finally, Park \textit{et al.} propose in \cite{Park-FingerprintPAD-tinyFCNN-TIFS-2019} an efficient CNN based on the fire module of the SqueezeNet to optimise the hardware and time requirements. Evaluated over the LivDet 2011 to 2015, the CNN outperforms for some datasets the work presented in \cite{chugh2018fingerprint}, at the same time reducing over 6 times the execution time. It should be though noted that the performance of this PAD method under more challenging scenarios with unknown attacks or sensors remains unknown.

To sum up, the main drawback of the aforementioned methods is their high dependency both on the PAI fabrication materials and the capture device. To tackle these issues, several approaches based on handcrafted features have been followed. On the one hand, Rattani \textit{et al.} proposed in \cite{Rattani-OpenSetPAD-TIFS-2015} an automatic adaptation of Weibull-calibrated support vector machines (SVMs). Over the LivDet 2011 database, the obtained equal error rates (EERs) oscillated between 20 and 30\% for the best configuration in the presence of unknown PAI species. On the other hand, Ding and Ross analysed an ensemble of one-class SVMs trained only on bona fide data in \cite{Ding-EnsembelOCSVM-PAD-WIFS-2016}, which lowered the error rates to 10-22\% over the same dataset. 

More recently, in an extension of \cite{chugh2018fingerprint}, Chugh and Jain identified in \cite{chugh-SpoofBusterGeneral-ICB-2019} a subset of six out of 12 PAI species which can yield detection rates similar to known attacks scenarios. That is, training the SpoofBuster with only those six PAI species and testing on all 12 species results in an APCER = 10.24\% at BPCER = 0.2\%, very close to the APCER = 9.03\% when all PAI species are used for training. In spite of these impressive results, it should be noted that the selection of the training PAI plays a crucial role in this study.

This dependecy is highlighted again by Engelsma and Jain in \cite{engelsma-oneClassGANsPAD-ICB-2019}, where multiple generative adversarial networks (GANs) are trained on bona fide images acquired with the RaspiReader sensor. From the same 12 different PAI species, six are used for training and six for testing. In a benchmark with the method proposed in \cite{Ding-EnsembelOCSVM-PAD-WIFS-2016}, the GANs outperform the SVMs. However, the average APCERs achieved for a BPCER = 0.2\% vary from 31.42\% to 68.98\%, depending on the training set used. This shows again a high sensitivity to different training datasets. In addition, this approach is not directly comparable to those based on conventional (e.g., Crossmatch or Greenbit) sensors, since a specific hardware, namely the RaspiReader, was used to acquire the samples.


Finally, Gajawada \textit{et al.} try to tackle this dependency on the PAI species contained in the training set from a different perspective in \cite{gajawada-PADCNNTranslator-ICB-2019}. They propose a so-called deep learning based \lq\lq Universal Material Translator'' (UMT). Given a reduced number (e.g., five) of samples from a new PAI species, the UMT extracts their main appearance features to embed them into a database of bona fide samples, in order to generate synthetic samples of the new PAI species. Those synthetic samples can be then utilised to train any CNN. Over the LivDet 2015 database, the authors showed how the proposed approach can improve up to 17\% the detection rates, achieving a remarkable 21.96\% APCER for a BPCER = 0.1\%. However, it should be noted that this approach does require some samples (i.e., five) of the analysed unknown PAI species.

In this context, our method tackles the issue of detection performance degradation in the presence of unknown factors (i.e., attacks, sensors, or databases) by transforming the local descriptors extracted from the fingerprint samples into a common feature space. This allows for better generalisation capabilities to more challenging scenarios, not needing any samples of the unknown attacks for training.



\section{Proposed method}
\label{sec:proposals}


Fig.~\ref{fig:General_overview} shows an overview of the proposed PAD approach, based on the fusion of three different feature encoding approachs. In the first common processing step, the Pyramid Histogram of Visual Words (PHOW) \cite{Bosch07a} algorithm is used to extract local features: the so-called dense Scale-Invariant Feature Transform (dense-SIFT) descriptors (Sect.~\ref{sec:DSIFT}). Subsequently, three encoding methods are applied to bring the aforementioned local descriptors into a common feature space: $i)$ Bag-of-Words (BoW)~\cite{Csurka04} (Sect.~\ref{sec:BoW}), \textit{ii)} Fisher Vector (FV)~\cite{sanchez2013image} (Sect.~\ref{sec:FV}), and \textit{iii)} Vector Locally Aggregated Descriptors (Vlad)~\cite{jegou2012aggregating} (Sect.~\ref{sec:Vlad}). Afterwards, each set of encoded features is classified using a different Support Vector Machine (SVM) (Sect.~\ref{sec:classif}). The final bona fide (BP) vs presentation attack (PA) decision for the sample at hand is defined as a weighted score level fusion of the three SVMs (Sect.~\ref{sec:fusion}).

\subsection{Local Features Extraction: dense-SIFT Descriptors}
\label{sec:DSIFT}

\begin{figure}[t]
\centering
\includegraphics[width = 0.85\linewidth]{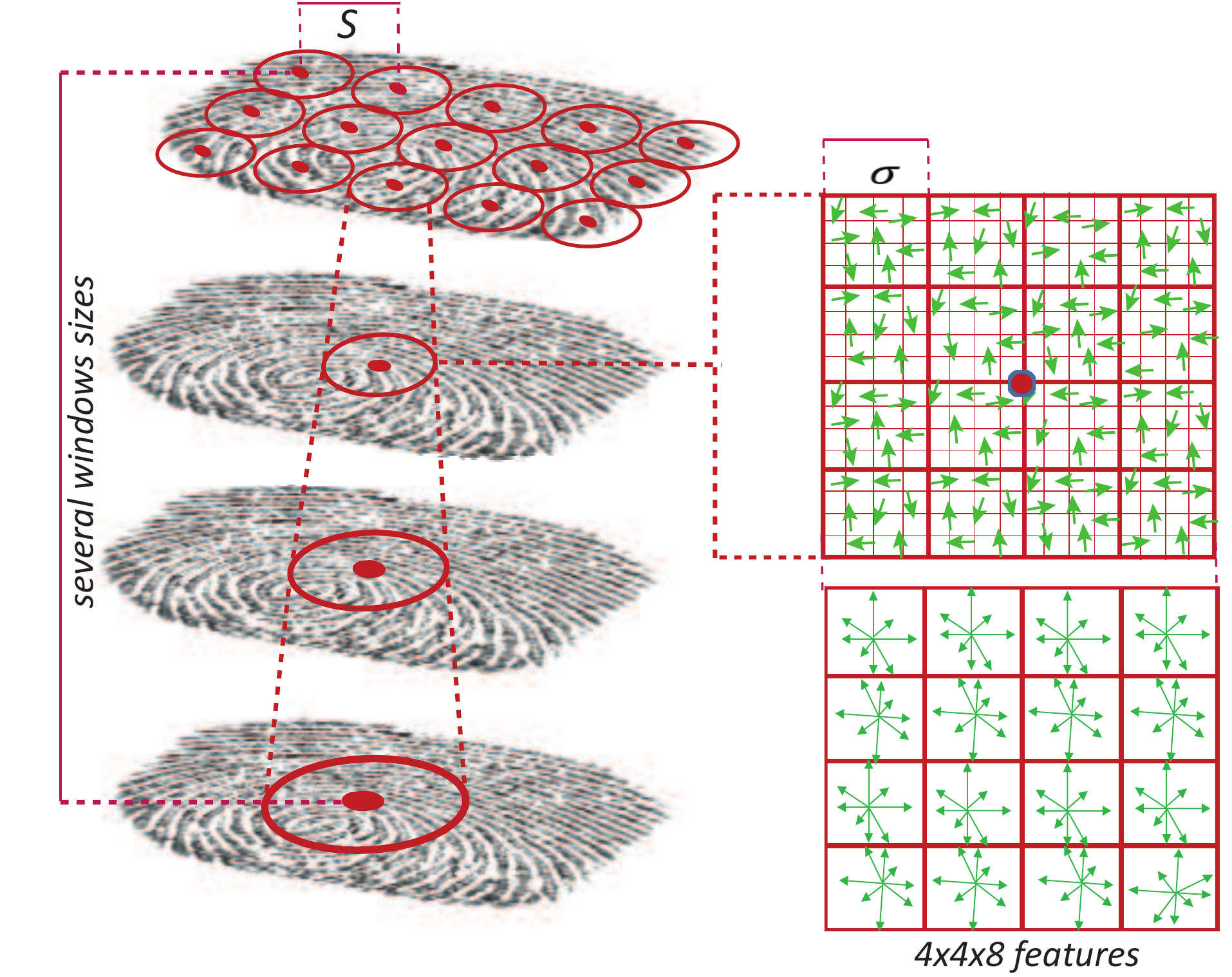}
\caption{dense-SIFT descriptors computed on fixed points on a regular grid, striding with a uniform spacing $S$ and using several scales $\sigma$.}
\label{fig:SIFT}\vspace*{-0.5cm}
\end{figure}

As local feature descriptors we have chosen the dense-SIFT approach, computed over the image gradient, since they can capture lower coherence areas introduced by the coarseness of different PAI fabrication materials. In particular, the Pyramid Histogram Of visual Words (PHOW) approach proposed by \cite{Bosch07a} computes SIFT descriptors densely at fixed points on a regular grid with uniform spacing $S$ (e.g., 5 pixels), as summarised in Fig.~\ref{fig:SIFT} (left). For each point in the grid, the dense-SIFT descriptor computes the gradient vector for each pixel in the feature point's neighbourhood (Fig.~\ref{fig:SIFT}, top right), taking into account 8 different directions. Subsequently, a normalized 8-bin histogram of gradient directions (Fig.~\ref{fig:SIFT}, bottom right) is built over $4 \times 4$ sample regions. In addition, in order to account for the scale variation between fingerprints, these dense-SIFT descriptors are computed over four circular patches or windows with different scales $\sigma = \left\lbrace 5, 7, 10, 12\right\rbrace$. Therefore, each point in the grid is represented by four SIFT descriptors (i.e., one per $\sigma$) comprising a total number of 128 features (i.e., $4\times4$ 8-bin histograms). 

It should be noted that windows with different scales allow extracting local information of fingerprints at different resolution levels, thereby detecting variable-size artefacts produced in the fabrication of PAIs. In addition, near-uniform local patches do not yield stable keypoints or descriptors. Therefore, we have used a fixed threshold $\delta$ on the average norm of the local gradient in order to remove local descriptors from low contrast regions (i.e., regions with an average norm value close to zero). 

\subsection{Local Feature Encoding}

In the second stage of the PAD algorithm, three different feature encoding approaches for the dense-SIFT descriptors are analysed. 

\vspace*{0.2cm}

\subsubsection{\textbf{Bag of Words (BoW)}}
\label{sec:BoW}

Bag-of-Words (BoW) based techniques were first developed for text categorization tasks, in which a text document is assigned to one or more categories based on its content \cite{lodhi2002text}. For this purpose, BoW represents the text document by a sparse histogram of word occurrence based on a visual vocabulary. Following this same idea, Csurka et al. \cite{Csurka04} adopted and transformed this technique to represent local features from an image in terms of the so-called \textit{visual words}. Our method builds upon this last approach.

\begin{figure}[t!]
\centering
\includegraphics[width = \linewidth]{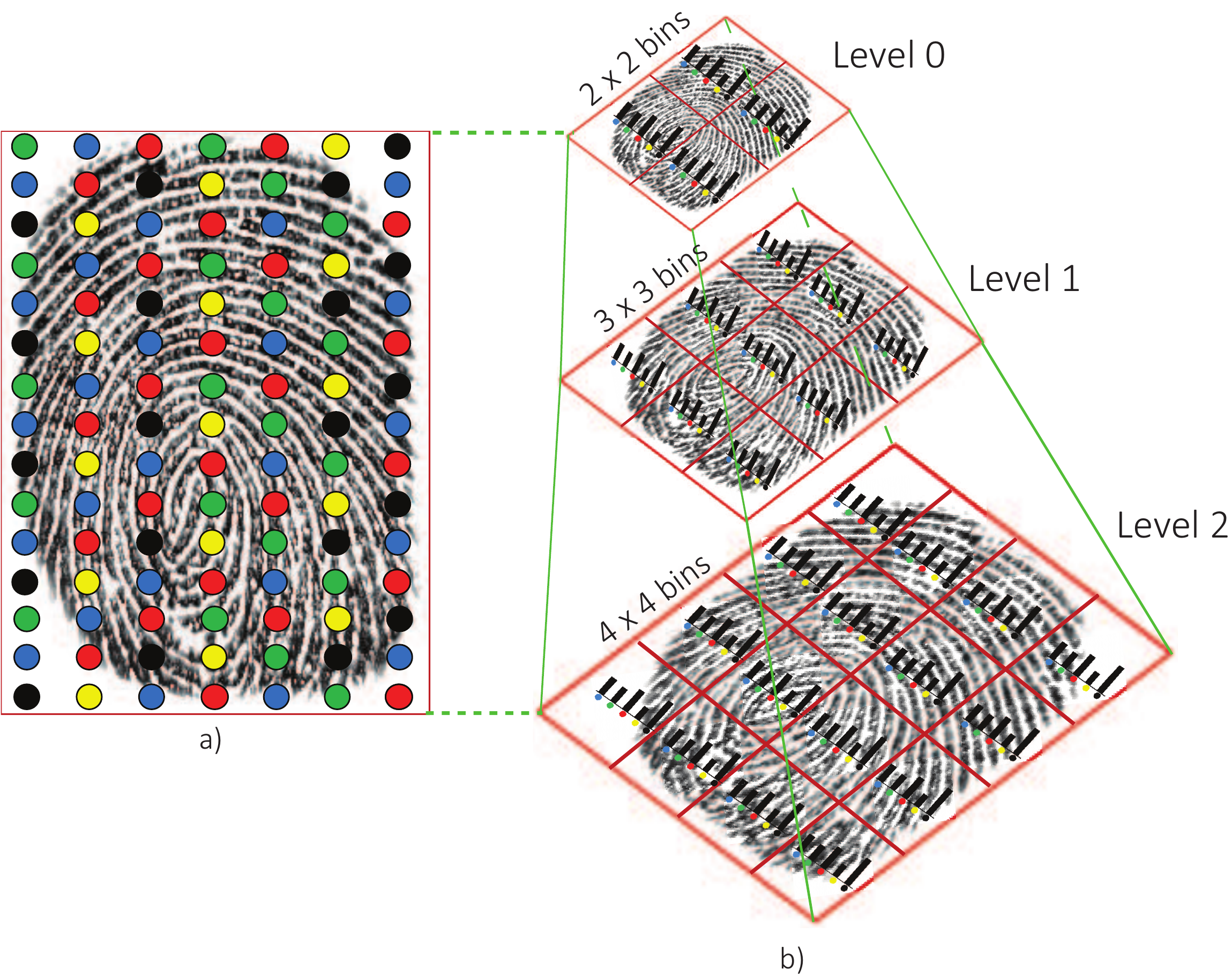}
\caption{Example of pyramid of spatial histograms. a) Quantized features using $k$-means. b) 3-level pyramid of spatial histograms built from quantized features.}
\label{fig:SpatialH}\vspace*{-0.5cm}
\end{figure}

As first proposed in \cite{janier17PADbagWords}, the BoW representation first computes the visual vocabulary as a codebook with $K$ different centroids or visual words (see Fig.~\ref{fig:General_overview}, top) with $k$-means clustering. Then, the BoW representation is defined as the histogram of the number of image descriptors assigned to each visual word. Its computation is summarised in Fig.~\ref{fig:SpatialH}. First, an $m$-level pyramid of spatial histograms is used in order to incorporate spatial relationships between patches. To do that, the fingerprint image is partitioned into increasingly fine sub-regions, and the dense-SIFT descriptors inside each sub-region are assigned to the closest centroid among the $K$ visual words, using a fast version of $k$-means clustering~\cite{elkan2003using}. Subsequently, the histograms inside each sub-region are computed and stacked into a single and final feature vector. 

\vspace*{0.2cm}

\subsubsection{\textbf{Fisher Vector (FV)}}
\label{sec:FV}

BoW approaches encode local features using a \textit{hard assignment}, in which a local descriptor is only assigned to one visual word based on a similarity function. In contrast, the Fisher Vector (FV) method derives a kernel from a generative model of the data (e.g., Gaussian Mixture Model, GMM), and describes how the set of local descriptors deviate from an average distribution of the descriptors \cite{sanchez2013image}. The aforementioned model can be understood as a \textit{probabilistic} visual vocabulary, which thereby allows a \textit{soft assignment}. Thus, the FV paradigm encodes not only the number of descriptors assigned to each region, but also their position in terms of their deviation with respect to the pre-defined model. 

As proposed in \cite{perronnin2010improving}, we train a GMM model with diagonal covariances from decorrelated dense-SIFT descriptors extracted on the previous step (see the second row in Fig.~\ref{fig:General_overview}). In general, the $K$-components of the GMM are represented by the mixture weights~($w_k$), Gaussian means~($\mu_k$) and covariance matrix~($\sigma_k$), with $k = 1, \dots, K$. This leads to an image representation which captures the average statistics first-order and second-order differences between the local features and each of the GMM centres~\cite{simonyan2013fisher}:

\begin{align}
	\phi_k^{1} = & \frac{1}{N\sqrt{w_k}}\sum_{i = 1}^N \alpha_i (k) \left(\frac{x_i - \mu_k}{\sigma_k}\right), \\	
	\phi_k^{2} = & \frac{1}{N\sqrt{2w_k}}\sum_{i = 1}^N \alpha_i (k) \left(\frac{(x_i - \mu_k)^2}{\sigma_k^2} - 1\right), 
\end{align}

\noindent where $\alpha_i(k)$ is the soft assignment weights of the $i$-th feature $x_i$ to the $k$-th Gaussian. It is important to highlight that $w_k, \mu_k$ and $\sigma_k$ are computed during the training stage. Finally, the FV representation that defines a fingerprint image is obtained by stacking the differences: $\phi = \left[ \phi_1^{1}, \phi_1^{2}, \cdots,  \phi_K^{1}, \phi_K^{2} \right]$.

\begin{table*}[t!]
\centering
\scriptsize
\caption{PAI fabrication materials used in each dataset of the LivDet 2011 - 2015 databases, where U denotes unknown material in the test set.}.
\label{tab:databases}
\begin{adjustbox}{max width=\textwidth}
\begin{tabular}{l l c c c c c c c c c c c c}                     
\toprule
DB & Dataset & Gelatine & Latex & PlayDoh & Silicone & Wood glue & Ecoflex & Body Double & Modasil & Liquid ecoflex & RTV & OOMOO & Gelatine 2\\
\midrule
 \multirow{4}{*}{\rotatebox{90}{2011}}          
 & Biometrika & \checkmark & \checkmark & & \checkmark &\checkmark &\checkmark & \\
 & Digital P.& \checkmark & \checkmark & \checkmark & \checkmark & \checkmark &\\
 & Italdata & \checkmark & \checkmark & & \checkmark &\checkmark &\checkmark & \\
 & Sagem    & \checkmark & \checkmark & \checkmark & \checkmark & \checkmark & \\   
 \midrule	
  \multirow{4}{*}{\rotatebox{90}{2013}}          
 & Biometrika & \checkmark  & \checkmark & & & \checkmark& \checkmark  &   & \checkmark \\
 & Italdata & \checkmark  & \checkmark & & & \checkmark & \checkmark  &   & \checkmark \\
 & Crossmatch &  & \checkmark & \checkmark & & \checkmark& & \checkmark & \\
 & Swipe    &  & \checkmark & \checkmark & & \checkmark& & \checkmark & \\
 \midrule	
  \multirow{4}{*}{\rotatebox{90}{2015}}          
 & GreenBit & \checkmark & \checkmark &  & & \checkmark & \checkmark &  & & \textbf{\textcolor{ForestGreen}{\checkmark (U)}} & \textbf{\textcolor{ForestGreen}{\checkmark (U)}}  &  & \\
 & Digital P. & \checkmark & \checkmark &  & & \checkmark & \checkmark &  & & \textbf{\textcolor{ForestGreen}{\checkmark (U)}} & \textbf{\textcolor{ForestGreen}{\checkmark (U)}}  &  & \\
 & Hi\_Scan & \checkmark & \checkmark &  & & \checkmark & \checkmark &  & & \textbf{\textcolor{ForestGreen}{\checkmark (U)}} & \textbf{\textcolor{ForestGreen}{\checkmark (U)}}  &  & \\
 & Crossmatch    & & & \checkmark & & & \checkmark & \checkmark & & & & \textbf{\textcolor{ForestGreen}{\checkmark (U)}} & \textbf{\textcolor{ForestGreen}{\checkmark (U)}} \\ 
 \bottomrule   		   
\end{tabular}
\end{adjustbox}
\end{table*}

With the aim of clustering the extracted local features with GMM diagonal covariance matrices, the dense-SIFT features are decorrelated using PCA~\cite{jegou2012aggregating}. In our approach, the dense-SIFT descriptor dimension was reduced from 128 to $d$ = 64 components, hence resulting the final FV representation in a $2Kd = 128\cdot K$ size vector, where $K$ is the number of Gaussian components in the GMM and $d$ is the dimension of a dense-SIFT descriptor.

\vspace*{0.2cm}

\subsubsection{\textbf{Vector Locally Aggregated Descriptors (Vlad)}}
\label{sec:Vlad}

In order to reduce the high-dimension image representation proposed by the FV and BoW approaches, gaining in efficiency and memory usage, we have finally studied the Vector Locally Aggregated Descriptors (Vlad) methodology \cite{jegou2012aggregating} (see Fig.~\ref{fig:General_overview}, third row). This is a simplified non-probabilistic version of FV, which models the data distribution from the accumulative distances between a visual word $\mathbf{x_i}$ and its closest center $\mathbf{c}$ in the visual vocabulary. Therefore, as in the BoW approach, a visual vocabulary needs to be computed in the first step with the $k$-means algorithm.

More specifically, a $d$-dimensional local feature descriptor $\mathbf{x}$ (i.e., dense-SIFT descriptor) can be represented by a Vlad descriptor $\mathbf{v_x}$ of size $Kd$ as follows:
\begin{equation}
\mathbf{v_x} = \sum_{j = 1}^d \left(\sum_{x:NN(x) = c_i} x_j - c_{i,j}\right),
\end{equation}   
where $x_j$ and $c_{i,j}$ denote the $j$-th component of $\mathbf{x}$, and its corresponding closest visual word $\mathbf{c_i}$. In our method, $\mathbf{v_x}$ is subsequently $L_2$-normalised in order to further improve the classification accuracy. 

Finally, it is important to highlight that Vlad also uses PCA for decorrelating training data.


\subsection{Classification}
\label{sec:classif}

In order to classify the final encoded representations, separate linear SVMs have been used for each encoding approach. In order to find the optimal hyperplane separating the bona fide from the attack presentations, the optimisation algorithm bounds the loss from below. Therefore, we have trained two complementary SVMs as follows:
\begin{itemize}
\item The first SVM labels the bona fide samples as +1 and the presentation attacks as -1, thereby yielding the corresponding $W_{bf}$ (weights) and $\mathbf{b}_{bf}$ (bias) classifier parameters.

\item The second SVM labels the bona fide samples as -1 and the presentation attacks as +1, thereby yielding the corresponding $W_{pa}$ and $\mathbf{b}_{pa}$ classifier parameters.
\end{itemize}

Subsequently, given an encoded feature descriptor $\mathbf{x}$, two different scores are computed, which estimate both the class of the sample (i.e., the score sign) and the confidence of such decision (i.e., the absolute value of the score is the distance to the hyperplane):
\begin{eqnarray}
s_{bf} &= W_{bf}'\cdot \mathbf{x} + \mathbf{b}_{bf}
\\
s_{pa} &= W_{pa}'\cdot \mathbf{x} + \mathbf{b}_{pa}
\end{eqnarray}
The final score is then computed to minimise the distance to the corresponding hyperplane, thereby choosing the most reliable decision for the given vector:
\begin{equation}
s = \begin{cases} 
s_{bf} &\mbox{if } \mathrm{abs}\left(s_{bf}\right) < \mathrm{abs}\left(s_{pa}\right) \\ 
s_{pa} & \mbox{otherwise} 
\end{cases}
\end{equation}

\subsection{Fused Approach (FPAD)}
\label{sec:fusion}

Given three different individual PAD scores, $s_\mathrm{FV}, s_\mathrm{Vlad}, s_\mathrm{BoW}$, output by the corresponding SVM, we define the final fused score $s_\mathrm{fusion}$ as follows:
\begin{equation}
s_\mathrm{fusion} = \alpha \cdot s_\mathrm{FV} + \beta \cdot s_\mathrm{Vlad} + \left( 1 - \alpha - \beta\right) \cdot s_\mathrm{BoW}
\end{equation}
where $\alpha + \beta \le 1$.

\section{Experimental evaluation}
\label{sec:Results}

In this section, we evaluate and benchmark the detection performance of each fingerprint encoding scheme described in Sect.~\ref{sec:proposals}. Specifically, three goals were taken into account for the experimental protocol design: $i)$ analyse the impact of the key parameter $K$ (vocabulary size) on the detection performance of the three proposed PAD schemes, $ii)$ benchmark the detection performance of our proposals against the top state-of-the-art approaches, and $iii)$ study the computational performance of the three fingerprint encoding schemes.

\subsection{Experimental Protocol}

The proposed PAD methods were implemented in C++ using the open-source VLFeat library\footnote{\url{http://www.vlfeat.org/}}. All the experiments were conducted on an Intel(R) Xeon(R) CPU E5-2670 v2 processor at 2.50 GHz, 378GB RAM. 

\subsubsection{\textbf{Databases}}

The experiments were conducted on the well-established benchmarks from LivDet 2011 \cite{yambay2012livdet}, LivDet 2013 \cite{ghiani2013livdet} and LivDet 2015 \cite{mura2015livdet}. A summary of the PAI fabrications materials is included in Table~\ref{tab:databases}.

\begin{figure*}[t]
	\centering
	\includegraphics[width=0.8\linewidth]{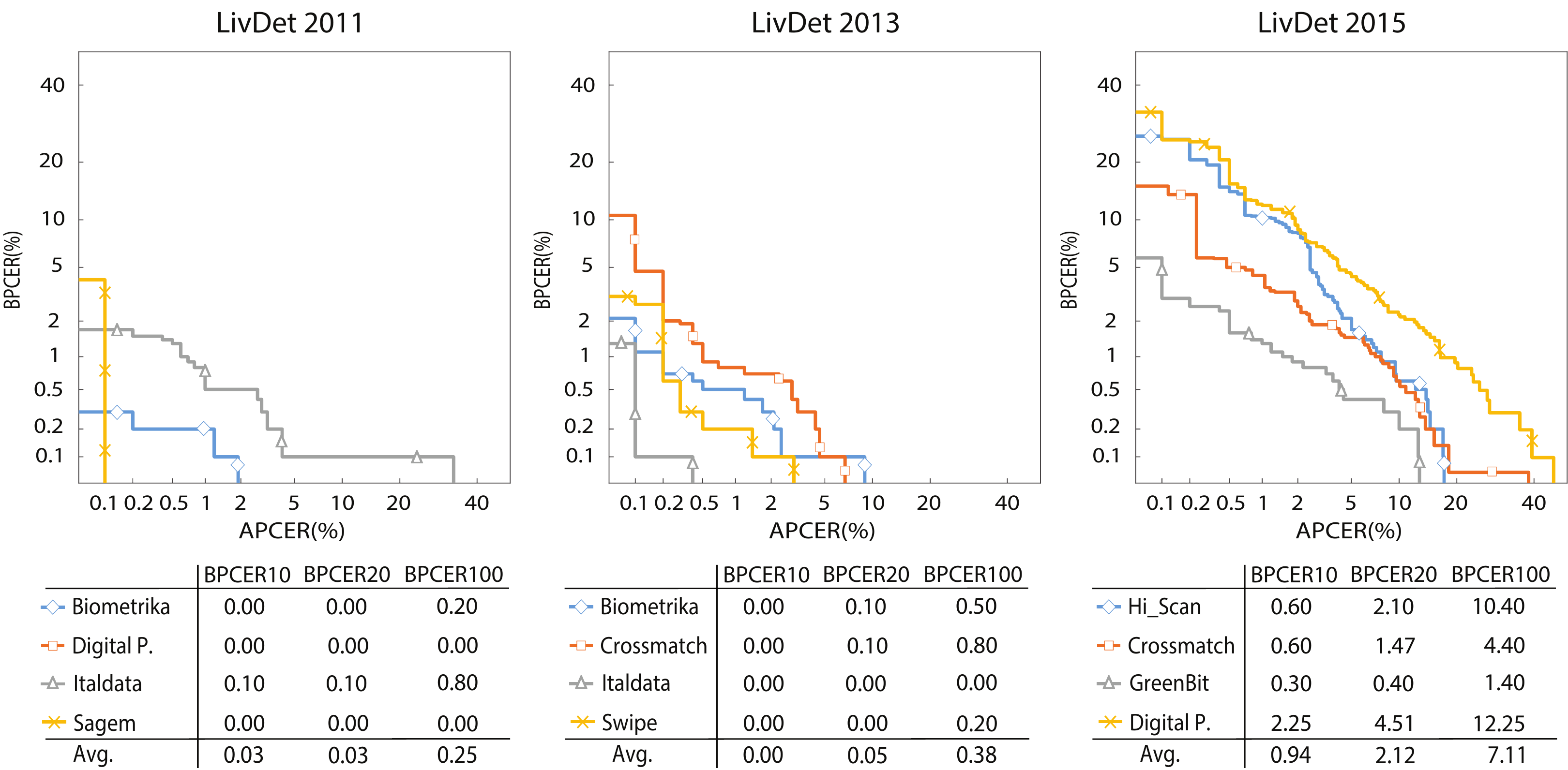}
	\caption{Performance evaluation of fused FPAD method under the \textbf{known-material and known-sensor} scenario.}\label{fig:DET_Known_fusion} \vspace*{-0.3cm}
\end{figure*}

\begin{table*}[t]
\centering
\caption{Benchmark in terms of the ACER with the state-of-the-art. The best results are highlighted in bold. }
\label{tab:comparison_SOTA}
\begin{tabular}{c l c c c c c c c c c } 
\toprule
DB & Dataset &  \cite{nogueira2016fingerprint} 
                      &  \cite{pala2017deep}        &  \cite{chugh2017fingerprint}  
                      & \cite{Park-FingerprintPAD-tinyFCNN-TIFS-2019}
                      &  FSB~\cite{chugh2018fingerprint}
                      &  FV         &   Vlad          &  BoW  & FPAD  \\

\midrule
\multirow{5}{*}{\rotatebox{90}{LivDet 2011}} 
& Biometrika    &  5.20  & 5.15   & 2.60  & 1.10 & 1.24 & 3.45  & 3.90 &  8.15   &  	\textbf{0.20} ($\alpha = 0.40, \beta = 0.40$)  \\ 

& Digital P.    &  3.20  & 1.85   & 2.70   & 1.10 &1.61 & 0.20  & 0.10 &  3.15	&	\textbf{0.00} ($\alpha = 0.00, \beta = 0.50$)\\

& Italdata      &  8.00  & 5.10   & 3.25   & 4.75 & 2.45 & 3.10  & 6.50 &  11.15	&	\textbf{0.80} ($\alpha = 0.50, \beta = 0.10$)\\ 

& Sagem         &  1.70  & 1.23   & 1.80   & 1.56 & 1.39 & 1.75  & 1.00 &  4.35	&	\textbf{0.10} ($\alpha = 0.00, \beta = 0.60$) \\

&   Avg.        &  4.52  & 3.33   & 2.59   & 3.12 & 1.67 & 2.13  & 2.88 &  6.70	&	\textbf{0.28}\\
\midrule
\multirow{5}{*}{\rotatebox{90}{LivDet 2013}} 
& Biometrika    &  1.80  & 0.65   & 0.60    & 0.35 & \textbf{0.20} & 1.30 & 1.70 &  4.95	&	0.50 ($\alpha = 0.10, \beta = 0.70$) \\ 

& Italdata      & 0.40  & 0.50    & 0.40       & 0.40 & 0.30 & 0.60 & 0.70 &  12.25 &	\textbf{0.10} ($\alpha = 0.20, \beta = 0.50$) \\

& Crossmatch    &  3.40  & -     & -         &  - &   -  &  3.70  & 4.30 &  5.50&	\textbf{0.80} ($\alpha = 0.60, \beta = 0.00$)\\ 

& Swipe         &  3.70  & 0.66    & -        & -     & -     & 1.90 & 4.00 &  6.35 & \textbf{0.30} ($\alpha = 0.50, \beta = 0.10$)\\

& Avg.	        & 2.33  & - & -           & -     & - & 1.88 & 2.68 &  7.26	&	\textbf{0.43}\\
\midrule
\multirow{5}{*}{\rotatebox{90}{LivDet 2015}} 
& GreenBit     &  4.60  &  -    & 2.00  &  \textbf{0.35}   & 0.68 & 1.30 & 2.40 &  7.05	&	1.20 
($\alpha = 0.90, \beta = 0.00$) \\ 

& Digital P.   &  5.64  &  -     & 1.76    & \textbf{1.09}& 1.12 & 4.75 & 5.20 &  14.10	&	4.60 ($\alpha = 0.80, \beta = 0.20$)\\

& Hi\_Scan     &  6.28  &  -     & \textbf{1.08}    &3.40 &  1.48 & 3.20 & 4.20 &  11.15	&	3.20 ($\alpha = 1.00, \beta = 0.00$)\\ 

& Crossmatch   & 1.90  &  -      & 0.81    & \textbf{0.20} & 0.64 & 3.56 & 4.85 &  10.38	&	2.28 ($\alpha = 0.90, \beta = 0.10$)\\

& 	Avg.       & 4.61  &  -      & 1.39  & 1.26  & \textbf{0.97} & 3.20 & 4.16 &  10.67	&	2.82\\
\bottomrule
\vspace*{-0.5cm}
\end{tabular}
\end{table*}

\subsubsection{\textbf{Evaluation Protocol and Metrics}}

To reach the aforementioned objectives, the experimental evaluation considers three different scenarios: $i)$ known-material and known-sensor, $ii)$ known-sensor and unknown-material, and $iii)$ unknown-sensor and cross-database. 

The detection performance is evaluated in compliance with the ISO/IEC IS 30107 \cite{ISO-IEC-30107-3-PAD-metrics-170227}: we report the Attack Presentation Classification Error Rate (APCER), which refers to the percentage of misclassified presentation attacks for a fixed threshold, and the Bona Fide Presentation Classification Error Rate (BPCER), which indicates the percentage of misclassified bona fide presentations. We also include the Detection Error Trade-Off (DET) curves between both error rates, as well as the BPCER for a fixed APCER of $10\%$ (BPCER10), $5\%$ (BPCER20) and $1\%$ (BPCER100).

Then, in order to establish a fair benchmark with the existing literature, we report the ACER as the average of the APCER and the BPCER for a fixed detection threshold $\delta$.


\subsection{Experimental Results}

\subsubsection{\textbf{Known-Material and Known-Sensor Scenario}}

First, we optimise the algorithms' detection performances in terms of the main key parameter: the visual vocabulary size $K$. To that end, we focus on the known scenario, in order to avoid a bias due to other variables. We test the following range of values: $K=\{256, 512, 1024, 2048\}$, since $K > 2048$ would yield too long feature vectors, not usable for real-time applications. We found that the best $K$ value on average is $K = 1024$ (for more details, the reader is referred to the appendix), and optimised the fusion parameters (see Sect.~\ref{sec:fusion}) for this value in terms of the D-EER. 

Fig.~\ref{fig:DET_Known_fusion} shows the DET curves for the FPAD approach over all sensors for $K = 1024$. As it can be observed, for low APCER values of 1\% (i.e., high security thresholds), the FPAD achieves a remarkable average BPCER100 = 0.25\% (vs\.  4.05\% in~\cite{chugh2018fingerprint}) for LivDet 2011 and 0.38\% for LivDet 2013. More in detail, for LivDet 2011, the Digital Persona and Sagem sensors report a BPCER = 0\% for any APCER $\geq$ 0.2\%. Regarding the LivDet 2013 database, the results are similar and for all sensors, and we observe a BPCER = 0\% for any APCER $\geq$ 10\%. In contrast, the FPAD suffers a detection performance decrease, with error rates multiplied by up to 42 times. More specifically, it shows a BPCER10 = 0.94\%, BPCER20 = 2.12\% and BPCER100 = 7.11\%.



 \begin{figure*}[t]
	\centering
		\begin{subfigure}[b]{0.70\linewidth}
			\centering 
			\includegraphics[width=0.99\linewidth]{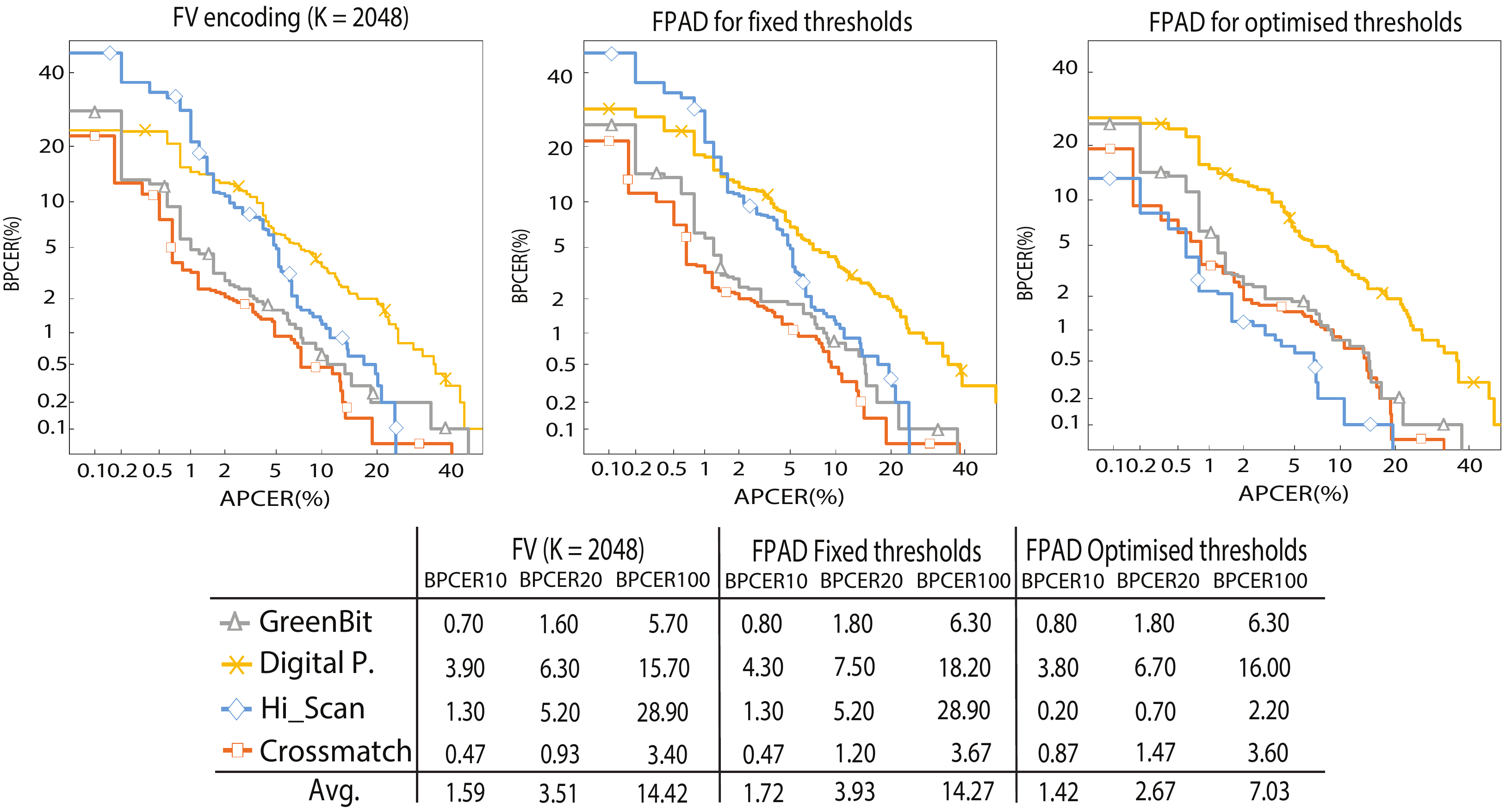}
			\caption{Unknown-material scenario from LivDet 2015.}\label{fig:ACE_livDet2015_unknown_DET}\vspace*{0.5cm}
		\end{subfigure}

		\begin{subfigure}[b]{0.70\linewidth}
			\centering
			\includegraphics[width=0.99\linewidth]{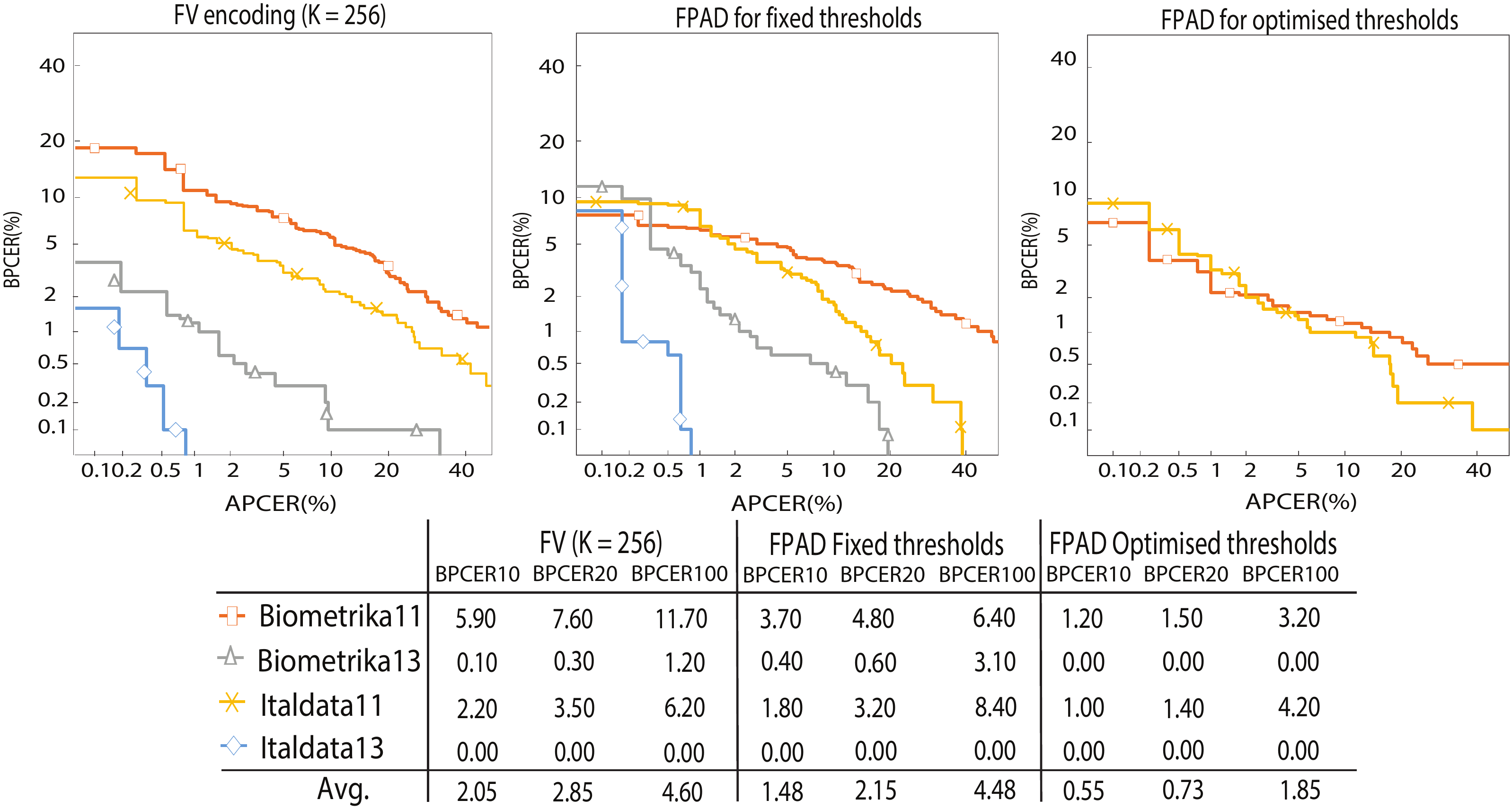}
			\caption{Unknown-material protocol from \cite{nogueira2016fingerprint}.}\label{fig:DET_UNKNOWN_NOGUEIRA_P}
		\end{subfigure}
	\caption{Performance evaluation over the \textbf{unknown material scenarios}.}\label{fig:DET_unknownSensors}
\end{figure*}

In Table~\ref{tab:comparison_SOTA}, we benchmark our results with the state-of-the-art in terms of the ACER. The lowest value on each row is highlighted in bold. As it can be observed, even if the individual feature encoding approaches do not outperform the FSB, the fused FPAD approach yields the lowest average ACER for both LivDet 2011 (0.28\% vs\.  1.67\%) and LivDet 2013 (0.43\%). On the other hand, the FSB achieves the best performance over LivDet 2015 (0.97\% vs\.  2.82\%). Nonetheless, it should be noted that the main goal of the present work is not only to achieve the best performance at a single operating point (i.e., the ACER is measured for $\delta = 0.5$) but overall for different applications requiring either a low BPCER (i.e., high convenience) or low APCER (i.e., high security), and also under more challenging and realistic conditions (i.e., unknown sensors or PAI species).

\subsubsection{\textbf{Known-Sensor and Unknown-Material Scenario}}
\label{sec:unkPAI}

In this scenario, both training and test samples were acquired by the same sensor, while presentation attacks in the test set were acquired from unknown PAI species. We analyse in detail the best performing single approach (FV) and the FPAD method. For the latter, we select the fixed thresholds obtained for the known-scenario (see $\alpha, \beta$ values in Table~\ref{tab:SOTA_ACE}), and denote this configuration as \lq\lq fixed thresholds''. In addition, we also evaluate its performance on the best $\alpha, \beta$ threshold combination (hereafter referred to as \lq\lq optimised thresholds''). The corresponding DET curves are reported in Fig.~\ref{fig:DET_unknownSensors}.


\begin{table*}[t]
\centering
\scriptsize
\caption{ACER evaluated on the \textbf{unknown-materials protocol} proposed by~\cite{nogueira2016fingerprint}.}\label{tab:cross_materials}

\begin{adjustbox}{max width=\linewidth}

\begin{tabular}{l c c | c c c | c c | c}                     
\toprule
                 	&\multicolumn{2}{c|}{\textbf{PAI species}}        & \multirow{2}{*}{\textbf{FV}} & \multirow{2}{*}{\textbf{Vlad}} &\multirow{2}{*}{\textbf{BoW}} & \multicolumn{2}{c|}{\textbf{FPAD}} & \multirow{2}{*}{\textbf{FSB}~\cite{chugh2018fingerprint}}
\\
\textbf{Dataset}  	&\textbf{Train}&\textbf{Test}  & & & &\textbf{Fx thr.}&\textbf{Op thr.} & \\
\hline
Bio11     & EcoFlex, Gelatine, Latex & Silgum, Woodglue    &  		6.33	& 10.05 & 15.05 &  4.78 & \textbf{2.05} & 4.60 \\
Bio13     & Modasil, Woodglue & EcoFlex, Gelatine, Latex &  1.00	& 2.82 & 5.50  &1.50 &\textbf{0.00} & 1.30  \\

Ita11      & EcoFlex, Gelatine, Latex & Silgum, Woodglue, Other &   4.50        & 16.50 & 21.83 & 3.60 & \textbf{2.00} & 5.20 \\
Ita13       & Modasil, Woodglue& EcoFlex, Gelatine, Latex   &  	0.50	   & 1.17 & 4.63 & 0.50 & \textbf{0.00} & \multirow{2}{*}{0.60} \\
 \midrule 
 	            &					&			 Avg.      &  3.08	        &    7.64    & 11.75         &   2.61 &  \textbf{1.01} &  2.93 \\
 \bottomrule   		   
\end{tabular}
\end{adjustbox}\vspace*{-0.3cm}

\end{table*}

Regarding the LivDet 2015 protocol, we can observe a similar behaviour between the FV encoding and the fused FPAD algorithm for fixed thresholds  in Fig.~\ref{fig:ACE_livDet2015_unknown_DET}. In particular, the BPCER10 and BPCER20 are slightly higher for the individual FV encoding (around 1.6-7\% and 3.5-9\%), but for high security thresholds, the FPAD achieves lower error rates (BPCER 14.3\% vs.\ 14.4\%). Also, the DET curves for Greenbit and Crossmatch are very close, whereas the performance for HI Scan and Digital Persona decreases. In contrast, the optimised thresholds FPAD achieves the best performance for Hi Scan, only showing a lower performance for Digital Persona. And in all cases, the detection rates are higher, yielding a low BPCER of 7\%. Regarding the state-of-the-art, \cite{gajawada-PADCNNTranslator-ICB-2019} achieves an average APCER of 22\% for a BPCER = 0.1\% for the Crossmatch dataset, and he FPAD approach achieves an APCER under 20\%, thus highlighting its soundness.


\begin{figure*}[t]
	\centering
		\begin{subfigure}[b]{0.70\linewidth}
			\centering 
			\includegraphics[width=0.99\linewidth]{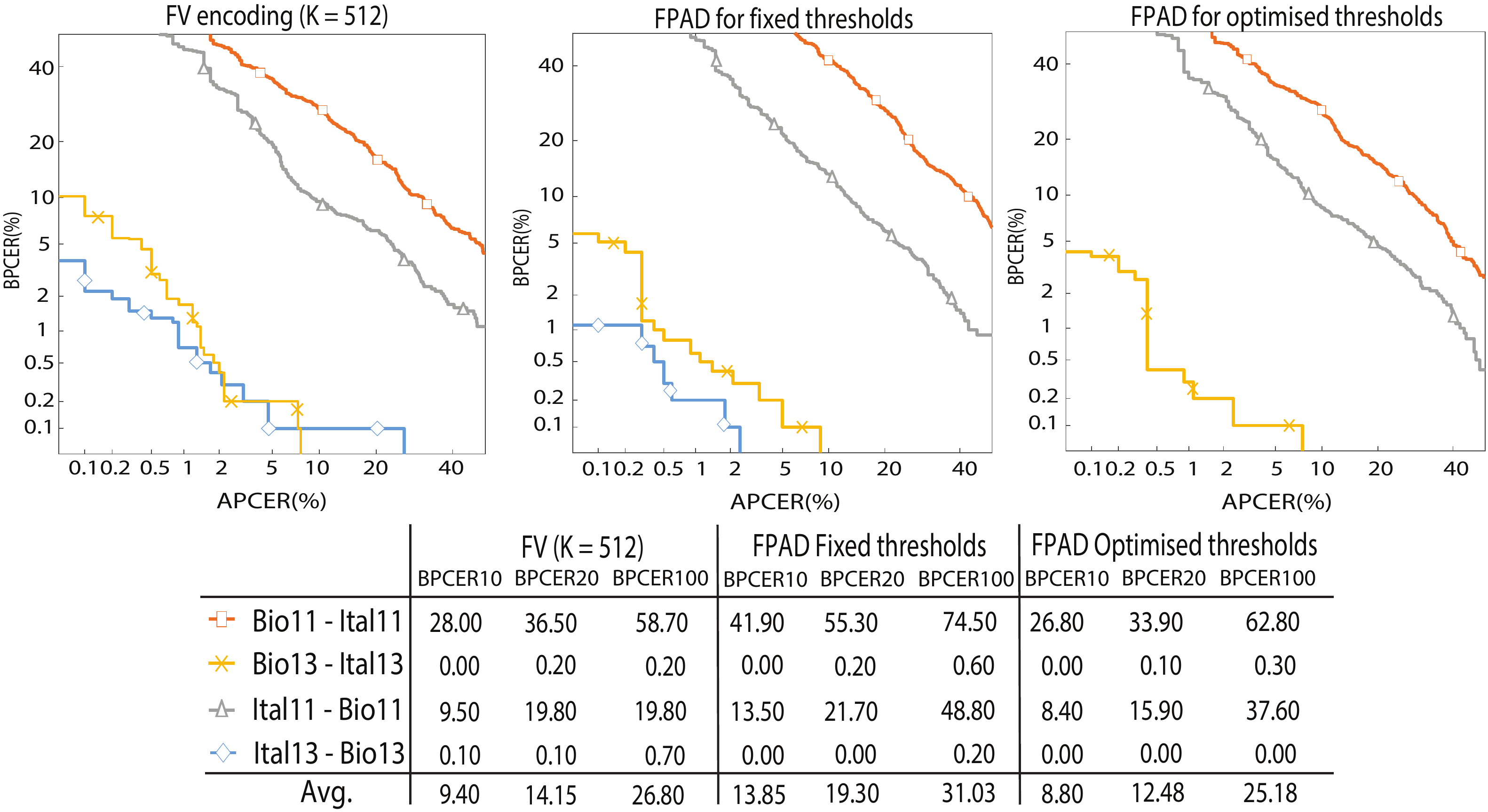}
			\caption{Unknown-sensor scenario.}\label{fig:cross-sensor_det}\vspace*{0.5cm}
		\end{subfigure}

		\begin{subfigure}[b]{0.70\linewidth}
			\centering
			\includegraphics[width=0.99\linewidth]{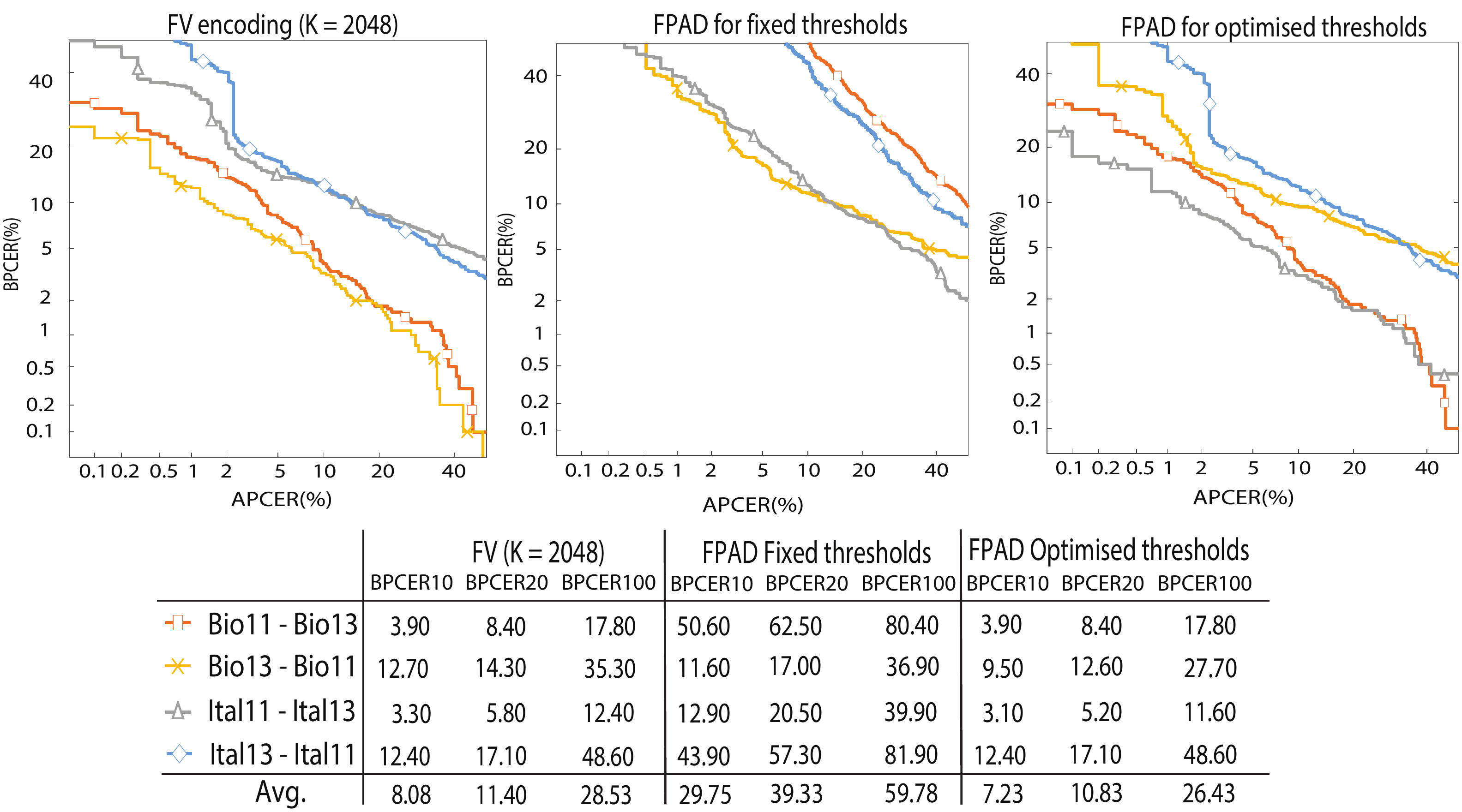}
			\caption{Cross-database scenario}\label{fig:cross-database}
		\end{subfigure}
	\caption{Performance evaluation over the \textbf{unknown sensor scenarios} proposed by \cite{nogueira2016fingerprint}.}\label{fig:DET_cross}\vspace*{-0.5cm}
\end{figure*}

In the second set of experiments, we follow the unkown-material protocol defined in \cite{nogueira2016fingerprint}. In this case, Fig.~\ref{fig:DET_UNKNOWN_NOGUEIRA_P} shows one of the main strengths of FV encoding: under high security scenarios, an average BPCER100 under 5\% can be achieved. In particular, for Italdata 2011 (BPCER100 = 6.20\%) and Italdata 2013 (BPCER100 = $0.0\%$) those values outperform the ones reported by \cite{chugh2018fingerprint}. Regarding the fused algorithms, it can be also observed that even the fixed thresholds configuration achieves a BPCER100 comparable to FSB~\cite{chugh2018fingerprint} (i.e., BPCER100 = 4.48\% vs.\ 4.24\%). In addition, the optimised thresholds FPAD reports a BPCER100 = 1.85\%, which is twice smaller.



We finally compare in Table~\ref{tab:cross_materials} the performance of our methods and FSB~\cite{chugh2018fingerprint} in terms of the ACER. We can observe that the FV encoding outperforms the remaining algorithms for three out of the four datasets. Moreover, for the fixed and optimised thresholds, our FPAD pipeline achieves an average ACER = 2.61\% and ACER = 1.01\% respectively, which considerably outperforms the top state-of-the-art.

 \vspace*{0.2cm}
\subsubsection{\textbf{Unknown-Sensor and Cross-Database Scenarios}}

Finally, we evaluate the soundness of our proposals in scenarios where different (i.e., unknown) sensors are used following the unknown-sensor and cross-database scenarios proposed by \cite{nogueira2016fingerprint}.

In the first set of experiments, training and test samples are acquired using different sensors (i.e., sensor inter-operability analysis). Fig.~\ref{fig:cross-sensor_det} shows the corresponding ISO-compliant evaluation. As it may be observed, training over the Italdata subset yields a better performance at all operating points than training over Biometrika (grey vs orange, and blue vs yellow cuves). Only low BPCERs $\le$ 0.5\% over the LivDet 2013 show a different behaviour. Moreover, for a fixed APCER of 1\%, the FV encoding achieves BPCER100 of 26.80\%, which reduces almost by 50\% the top state-of-the-art result (BPCER100 = 52.52\%) \cite{chugh2018fingerprint}. In addition, our optimised thresholds FPAD approach attains a BPCER = 0\% for all APCERs over the Italdata13 train set -- we may thus conclude that the method found the optimal common feature space from the Italdata 2013 training set to correctly classify the Biometrika 2013 samples.


Table~\ref{tab:cross_sensor} benchmarks all methods to FSB~\cite{chugh2018fingerprint} in terms of ACER. In general, and regardless of the particular train-test combination, FV encoding is able to outperform both the other two encoding approaches and the results obtained in \cite{chugh2018fingerprint} (i.e., average ACER = 7.83\% for FV vs.\ 14.59\% for FSB, which implies a relative improvement of 48\%). Moreover, the FPAD also outperforms the FSB~\cite{chugh2018fingerprint} for both the fixed and the optimised thresholds by a relative improvement of 38\% and 55\%, respectively. 

\begin{table}[t]
\centering
\scriptsize
\caption{ACER evaluated on the \textbf{unknown-sensor protocols} proposed by~\cite{nogueira2016fingerprint}.}

\begin{subtable}[t]{0.99\linewidth}
\centering
\caption{Unknown-sensor protocol.}
\label{tab:cross_sensor}
\begin{tabular}{c | c c c | c c | c }                     
\toprule
& \multirow{2}{*}{\textbf{FV}} & \multirow{2}{*}{\textbf{Vlad}} & \multirow{2}{*}{\textbf{BoW}}& \multicolumn{2}{c|}{\textbf{FPAD}}  & \multirow{2}{*}{\textbf{FSB}~\cite{chugh2018fingerprint}} \\
\textbf{Train - Test}  & & &  &\textbf{Fx thr.}&\textbf{Op thr.} &  \\
\midrule
Bio11 - Ital11       	 &    19.10 & 19.30 & 32.92 & 23.50  & \textbf{16.80}     &  25.35      \\
Bio13 - Ital13      		 &      1.70 & 3.50 & 32.20& 0.80  & \textbf{0.40}     &   4.30     \\
Ital11 - Bio11      		 &      9.60 & 15.20 & 32.25 &  11.40 & \textbf{9.10}    &  25.21     \\
Ital13 - Bio13      		 &       0.90 & 1.90 & 6.75  &	0.50 & \textbf{0.00}   &   3.50    \\ 
\midrule
Avg.                                 &      7.83 & 9.98 & 26.04   & 9.05 & \textbf{6.58} & 14.59 \\
\bottomrule
\end{tabular}
\end{subtable}

\vspace*{0.2cm}
\begin{subtable}[t]{0.99\linewidth}
\centering
\caption{Cross-database protocol.}\label{tab:cross_database}
\begin{tabular}{c | c c c | c c | c}                     
\toprule
& \multirow{2}{*}{\textbf{FV}} & \multirow{2}{*}{\textbf{Vlad}} & \multirow{2}{*}{\textbf{BoW}}& \multicolumn{2}{c|}{\textbf{FPAD}}  & \multirow{2}{*}{\textbf{FSB}~\cite{chugh2018fingerprint}} \\
\textbf{Train - Test}  & & &  &\textbf{Fx thr.}&\textbf{Op thr.} &  \\
\midrule
Bio11 - Bio13   &      \textbf{6.80} &     15.70 & 28.80 & 25.20 &  \textbf{6.80}    &  7.60 \\
Bio13 - Bio11   &     12.70 & 11.10 & 49.80 &	11.20 &	\textbf{9.50}     & 31.16 \\
Ital11 - Ital13       &      5.60 & 8.70 & 47.75   & 11.70	&	\textbf{5.10}     &  6.70\\
Ital13 - Ital11       &     \textbf{11.50} & 18.10 & 49.60&	22.90	&	\textbf{11.50}      & 26.16 \\ 
\midrule
Avg.                          &     9.15 & 13.40& 43.99 &	17.75	&	\textbf{8.23}    & 17.91 \\
\bottomrule
\end{tabular}
\end{subtable}\vspace*{-0.5cm}
\end{table}

In the second experiment, the performance is evaluated over the change of data collection over the same sensor (i.e., train and test over the same sensor, but acquired for LivDet 2011 and LivDet 2013, respectively). We refer to this protocol as cross-database scenario. In Fig.~\ref{fig:cross-database} we can see different behaviours for each algorithm for the different datasets. Whereas the Biometrika curves (orange and yellow) are very close for the FV encoding, this is not the case for the fused FPAD. This is due to the different generalisation capabilities of the remaining encoding approaches (BoW and Vlad), as it may be seen in Table~\ref{tab:cross_database}. In particular, the ACER achieved training over Biometrika 2011 are better than training over Biometrika 2013 for BoW (28.8\% vs.\ 15.70\%), and vice versa for Vlad (15/70\% vs.\ 11.10\%). In addition, the poor performance of BoW also affects the fixed thresholds FPAD, thereby yielding a poor BPCER100 of almost 60\%. However, the optimised thresholds FPAD can improve the error rates yielded by FV, achieving an average BPCER100 of 26\%. 

Finally, coming back to the ACER-based benchmark with FSB~\cite{chugh2018fingerprint}, we may observe that, on average, all the FV approach (ACER = 9.15\%), the fixed thresholds FPAD (ACER = 17.75\%) and the optimised thresholds FPAD (ACER = 8.23\%) are able to outperform the FSB (ACER = 17.91\%) by up to a 55\% relative improvement.

\vspace*{0.2cm}

\subsubsection{\textbf{Computational efficiency}}

In this last set of experiments, we study the computational efficiency of the proposed image encodings for different parameter configurations. For this purpose, we select the LivDet 2015 database, which contains the largest images. We found that the BoW encoding requires 0.38 seconds, Vlad 1.58 seconds, and FV 2.11 seconds. There is thus a trade-off between detection performance and time efficiency. However, in all cases, the algorithms can be utilised for real-time applications.

\section{Conclusions}
\label{sec:conclusions}

In this paper, we have proposed a new PAD method based on the combination of local dense-SIFT image descriptors and three different feature encoding approaches (i.e., FV, Vlad, and BoW). The experimental evaluation conducted over the publicly available LivDet 2011, LivDet 2013 and LivDet 2015 databases assessed the performance of our proposals with respect to the top state-of-the-art methods. The analysis of the detection performance showed that the FV reached the best individual detection accuracy for all databases. However, a score-level fusion of the three encoding approaches (known as FPAD) yielded an improved performance, significantly outperforming the top state-of-the-art results in the analysed scenarios, specially under the most challenging and realistic scenarios, where both unknown materials and unknown sensors are frequently employed. In addition, this fused approach achieved the highest detection accuracy on the LivDet 2019 competition \cite{Orru-LivDet2019-arxiv-2019}.

It should be also noted that the fixed thresholds configurations do not always outperform the FV encoding as a standalone algorithm. This highlights the challenges faced when unknown sensors or PAI species are contained in the test set. However, a proper tuning of the thresholds yields a very promising performance for the FPAD algorithm.


In more details, the ISO-compliant evaluation in terms of BPCER and APCER showed one of the main strengths of the FV encoding and the FPAD proposal: the low BPCERs achieved even for very high security operating points (i.e., APCER $\le$ 1\%). Specifically, the FPAD technique yielded an average BPCER100 of 25\% on the unkown-sensor scenario, and a BPCER100 of 26\% to 28\% on the cross-database scenario, thereby outperforming the top state-of-the-art results~\cite{chugh2018fingerprint} by up to a relative 50\% to 60\%, respectively. Moreover, both methods proved to be suitable in the presence of unknown PAI species, achieving a BPCER100 as low as 4.6\% and 1\%. In summary, the previous results indicate that $i)$ orientation histograms provided by the dense-SIFT method correctly represent the lack of continuity in the ridge's flow, and hence the artefacts produced in the fabrication of PAIs, and $ii)$ FV as well as the fusion-based proposal in combination with dense-SIFT descriptors found a new common feature space, which allows successfully detecting both known and unknown PAIs.

Finally, the computational efficiency evaluation showed that BoW encoding attained efficiency results below 400 milliseconds, while Vlad and FV encodings were above 1150 milliseconds. As future work lines, we will improve the computational cost of the Vlad and FV encodings in order to obtain the best trade-off between detection accuracy and computational efficiency.


\appendix

\section*{Analysis of the Detection Performance for Different Vocabulary Sizes}
\label{sec:ResultsK}

As it was mentioned in the article, the main parameter shared by all feature encoding approaches is the vocabulary size $K$. The larger $K$ is, the higher number of visual words is, and thus, the less the information loss during the quantisation carried out to convert the local dense-SIFT descriptors into the so-called common feature space. However, this also entails a higher computational cost, and can eventually end up in over fitting. Therefore, we analyse here in detail the impact of $K$ on the detection performance and the computational efficiency of the PAD method for each scenario.

\begin{figure*}[t]
	\begin{subfigure}[c]{0.99\linewidth}
		\centering
		\includegraphics[width=0.65\linewidth]{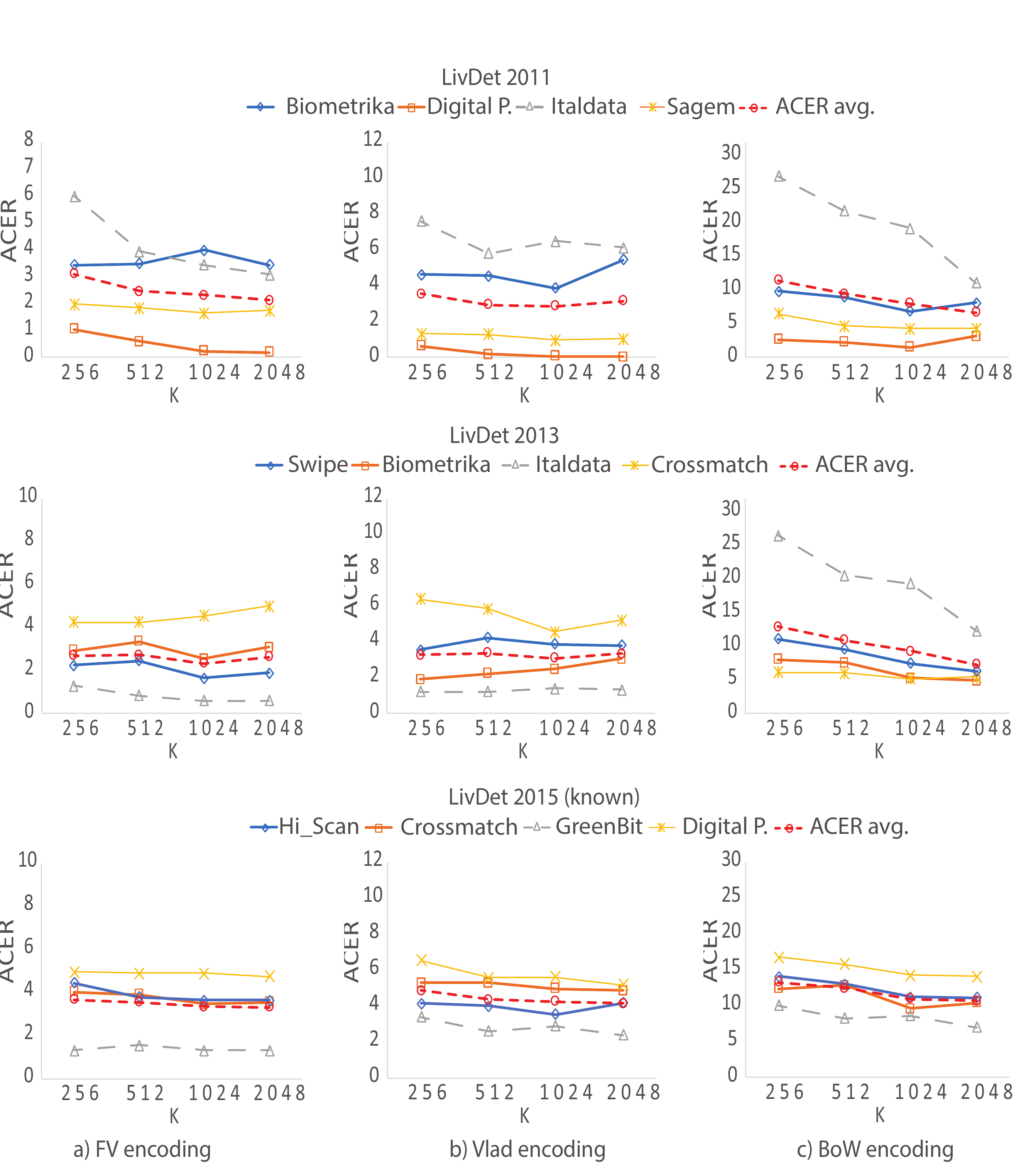}
		\caption{Known scenarios.}
		\label{fig:ACE_livDet_known_ACE} 
	\end{subfigure}
	
	\begin{subfigure}[c]{0.99\linewidth}
		\centering
		\includegraphics[width=0.7\linewidth]{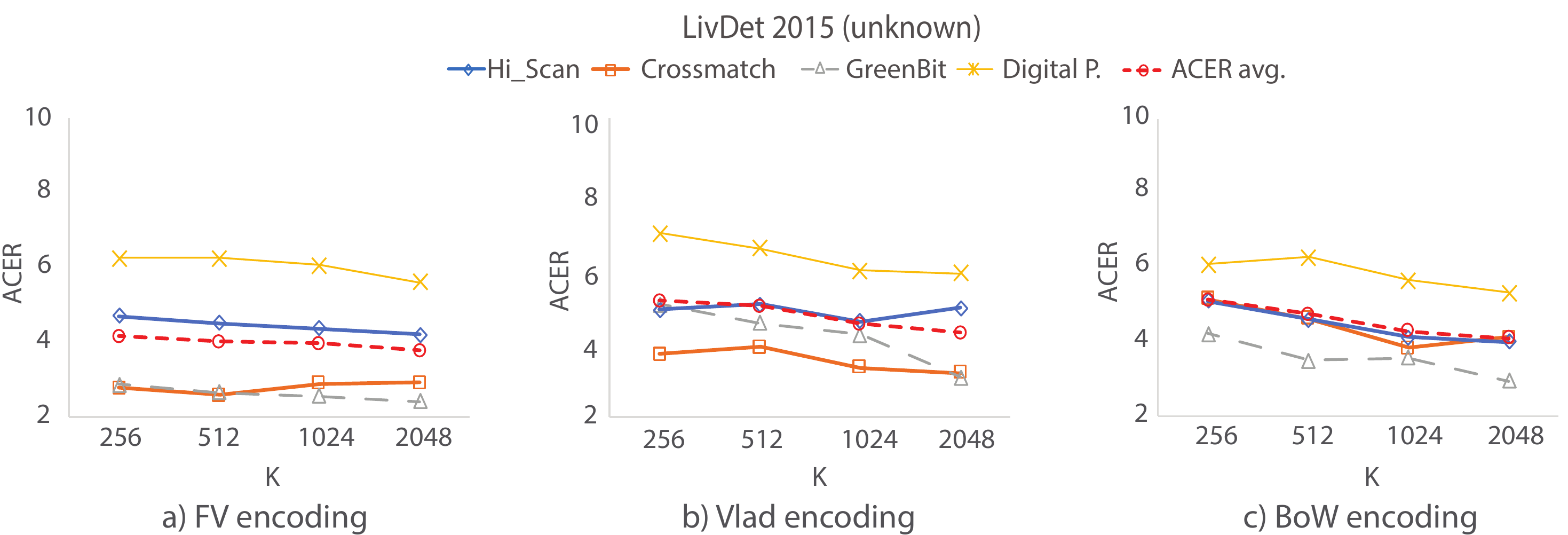}
		\caption{LivDet Unknown materials scenario}\label{fig:ACE_livDet2015_unknown_ACE} 
	\end{subfigure}
	\caption{Performance evaluation in terms of ACER for all databases. The dashed red line shows the average trend for different values of $K$.}\label{fig:acer}\vspace*{-0.5cm}
\end{figure*}

\subsection{\textbf{Known-Material and Known-Sensor Scenario}}

In the first place, we need to analyse the impact of  $K$ on the performance of the three proposed schemes individually. We do that under this all-known scenario in order to avoid a bias due to other variables (i.e., unknown PAI species or sensors). More specifically, we test the following range of values: $K=\{256, 512, 1024, 2048\}$, since $K > 2048$ would yield too long feature vectors, not usable for real-time applications. 

The ACER values for each method and $K$ are presented in Table~\ref{tab:comparison_SOTA}, and graphically in Fig.~\ref{fig:ACE_livDet_known_ACE}. As it can be observed, most curves reach a minimum (i.e., lowest ACER, and thus, best detection performance) for $K$ = 1024. In some cases, the ACER achieved for $K$ = 2048 continues to decrease (e.g., the BoW encoding for LivDet 2013), thus not reaching a minimum over the selected range. However, as it was mentioned above, such vocabulary sizes would imply a non real-time detection, and will thus not be considered in the present study.

Now, focusing on the best $K$ value on average, $K$ = 1024, we can highlight that FV encoding achieves on average, for all sensors, an ACER of 2.13\%, 1.88\% and 3.31\% on LivDet 2011, LivDet 2013 and LivDet 2015, respectively. On the other hand, the best Vlad performance values are found at $K$ = 1024 (i.e. 2.88\% on LivDet 2011 and 2.68\% on LivDet 2013) for all databases with exception of the LivDet 2015 dataset, in which the best accuracy is reached at $K$ = 2048 (ACER = 4.16\%). Finally, the BoW encoding improves its detection performance with $K$, thereby achieving its minimum ACER result at $K$ = 2048.

\subsection{\textbf{Known-Sensor and Unknown-Material Scenario}}

In this scenario, both training and test samples were acquired by the same sensor, while presentation attacks in the test set were acquired from unknown PAI species. 

In the first set of experiments, we select the LivDet 2015 database, since it already includes unknown PAI species for testing. Fig.~\ref{fig:ACE_livDet2015_unknown_ACE} shows, in terms of ACER, the impact of the parameter key $K$ on the performance of the proposed encoding techniques. As it can be seen, the average performance (represented with a dashed red line) improves with increasing values for $K$, achieving a minimum for $K$ = 2048. More specifically, the FV encoding yields the best ACER results, with an average value of $3.31\%$. 

We have also analysed the unknown materials protocols for LivDet 2011 and 2013 proposed in \cite{nogueira2016fingerprint}. The results are presented in Table~\ref{tab:cross_materials}. In this case, only the BoW encoding reaches the best detection performance for $K$ = 2048. On the other hand, on average, the best results are yielded by $K$ = 512 for FV, and $K$ = 256 for Vlad. 

Finally, it should also be highlighted that, for all three datasets (i.e., LivDet 2011, 2013 and 2015), BoW shows a higher variability range for different values of $K$. For instance, ACER varies within 3.03 and 4.44 for FV, between 1.64 and 8.61 for Vlad, and between 9.28 and 16.60 for BoW for LivDet 2015. Therefore, BoW is much more sensitive to changes in $K$.

\subsection{\textbf{Unknown-Sensor and Cross-Database Scenarios}}

Finally, in order to evaluate the soundness of our proposals in scenarios where different (i.e., unknown) sensors are used, we follow the unknown-sensor and cross-database scenarios proposed by \cite{nogueira2016fingerprint}.

In the first set of experiments, training and test samples are acquired using different sensors. Table~\ref{tab:cross_sensor} shows the ACER for different values of $K$. As it can be observed, the FV encoding achieves its better results at different values of $K$, depending on the sensor used for training: whereas for Italdata 2011 and 2013, the lowest ACER is achieved for $K$ = 512 (9.60\% and 0.90\%), for Biometrika it is obtained for $K$ = 2048 (18.50\% and 1.20\%). In general, and regardless of the particular train-test combination, FV encoding is able to outperform both the other two encoding approaches and the results obtained in \cite{chugh2018fingerprint} (i.e., ACER = 7.83\% for FV vs 14.59\% for FSB~ \cite{chugh2018fingerprint}, which implies a relative improvement of 48\%). These results indicate that FV encoding found a set of common features in training images that allow a correct detection of PAIs acquired with other sensors. 

\begin{table*}[t]
	\centering
	\caption{ACER evaluated on the known scenario and on the protocols proposed by~\cite{nogueira2016fingerprint}.}
	
	\begin{subtable}[t]{0.99\textwidth}
		\centering
		\caption{Known sensors and materials.}
		\label{tab:comparison_SOTA}
		\begin{tabular}{c c| c c c c |c c c c |c c c c} 
			\toprule
			&  & \multicolumn{4}{c|}{\textbf{FV encoding}} & \multicolumn{4}{c|}{\textbf{Vlad encoding}} & \multicolumn{4}{c}{\textbf{BoW encoding}}  \\
			\textbf{DB}  & \textbf{Dataset} &  \textbf{256} & \textbf{512} & \textbf{1024} & \textbf{2048}  &  \textbf{256} & \textbf{512} & \textbf{1024} & \textbf{2048} &  \textbf{256} & \textbf{512} & \textbf{1024} & \textbf{2048} 	\\
			\midrule
			\multirow{5}{*}{\rotatebox{90}{LivDet 2011}} 
			& Biometrika    &  \textbf{3.45}  & 3.50   & 4.00   & \textbf{3.45} & 4.65  & 4.55 &  \textbf{3.90}   & 5.45 &	9.85&	9.05 &	\textbf{6.95}	&	8.15	   \\ 
			
			& Digital P.    &  1.05  & 0.60   & 0.25   & \textbf{0.20} & 0.65  & 0.20 &  0.10	& \textbf{0.05} & 2.75&	2.30 &	\textbf{1.60}	&	3.15	  \\
			
			& Italdata      &  6.00  & 3.95   & 3.45   & \textbf{3.10} & 7.60  & \textbf{5.80} &  6.50	& 6.15 & 26.95 &	21.80 &	19.25	&	\textbf{11.15}	  \\ 
			
			& Sagem         &  2.00  & 1.85   & \textbf{1.65}   & 1.75 & 1.35  & 1.30 &  \textbf{1.00}	& 1.05 & 6.55&	4.70	&	\textbf{4.35}	&	\textbf{4.35}	  \\
			
			&   Avg.        &  3.13  & 2.48   & 2.34   & \textbf{2.13} & 3.56  & 2.96 &  \textbf{2.88}	& 3.18 & 11.53&	9.46 &	8.04	&	\textbf{6.70}	  \\
			\midrule
			\multirow{5}{*}{\rotatebox{90}{LivDet 2013}} 
			& Biometrika    &  2.95  & 3.35  & \textbf{2.55}  & 3.10  & \textbf{1.90} & 2.25 &  2.50	& 3.05 & 8.05  & 7.60 & 5.30 & \textbf{4.95}   \\ 
			
			& Italdata      & 1.30   & 0.85  & \textbf{0.60}  & 5.00  & \textbf{1.25} & \textbf{1.25} &  1.40  & 1.35 & 26.45 & 20.60 & 19.30 & \textbf{12.25}  \\
			
			& Crossmatch    &  4.25  & 4.25  & 4.55  & \textbf{0.60}  & 6.40 & 5.85 &  \textbf{4.55}  & 5.20 & 6.15  & 6.05 & \textbf{5.20} & 5.50  \\ 
			
			& Swipe         &  2.25  & 2.45  & \textbf{1.65}  & 1.90  & 3.55 & 4.25 &  3.90  & \textbf{3.80} & 11.15 & 9.60 	& 7.50 & \textbf{6.35}  \\
			
			& Avg.	        & 2.69   & 2.73  & \textbf{2.34}  & 2.65	 & 3.28 & 3.40 &  \textbf{3.09}	& 3.35 & 12.95 & 10.96 & 9.33 & \textbf{7.26}  \\
			\midrule
			\multirow{5}{*}{\rotatebox{90}{LivDet 2015}} 
			& GreenBit     &  \textbf{1.30}  &  1.55    & \textbf{1.30}    & \textbf{1.30} & 3.40 & 2.60 &  2.90	& \textbf{2.40} & 10.05 & 8.25 & 8.65 & \textbf{7.05}  \\ 
			
			& Digital P.   &  4.95  &  4.90    & 4.90    & \textbf{4.75} & 6.55 & 5.57 &  5.60	& \textbf{5.20} & 16.75 & 15.75 & 14.25 & \textbf{14.10}   \\
			
			& Hi\_Scan     &  4.45  &  3.80    & 3.65    & \textbf{3.20} & 4.15 & 4.00 &  \textbf{3.55}	& 4.20 & 14.05 & 13.05 & 11.20 & \textbf{11.15}  \\ 
			
			& Crossmatch   & 4.02  &  3.91     & \textbf{3.45}    & 3.56 & 5.31 & 5.30 &  4.97	& \textbf{4.85} & 12.41 & 12.83 & \textbf{9.58} & 10.38   \\
			
			& 	Avg.       & 3.68  &  3.54     & 3.33    & \textbf{3.20} & 4.85 & 4.37 &  4.26	& \textbf{4.16} & 13.32 & 12.47 & 10.92 & \textbf{10.67} \\
			\bottomrule
		\end{tabular}
	\end{subtable}
	
	\vspace*{0.2cm}
	\begin{subtable}[t]{0.99\textwidth}
		\centering
		\caption{Unknown-material protocol.}
		\label{tab:cross_materials}
		\begin{tabular}{c | c c c c |c c c c |c c c c}                  
			\toprule
			& \multicolumn{4}{c|}{\textbf{FV encoding}} & \multicolumn{4}{c|}{\textbf{Vlad encoding}} & \multicolumn{4}{c}{\textbf{BoW encoding}} \\
			\textbf{Dataset}  &  \textbf{256} & \textbf{512} & \textbf{1024} & \textbf{2048}  &  \textbf{256} & \textbf{512} & \textbf{1024} & \textbf{2048} &  \textbf{256} & \textbf{512} & \textbf{1024} & \textbf{2048}\\
			\midrule
			Bio11    & 		\textbf{6.33}	& 7.05	        &  8.33	   & 8.33 &  10.05 & \textbf{7.83}     &  9.05   & 10.05   &  17.98       & 15.50   & 15.05    & \textbf{11.83} \\
			Bio13     & \textbf{1.00}	& \textbf{1.00} &  1.18  & 1.32     & \textbf{2.82} &  3.13  & 4.13    &  3.82   & 8.70    &  6.05         &  5.50    &  \textbf{3.87}    \\
			Ita11        &   4.50	        & \textbf{3.78}	&  6.50	   & 7.78 & \textbf{16.50}  & 19.33     & 21.23   & 19.78   &  31.53        & 27.50    & 21.83        & \textbf{18.00}  \\
			Ita13      &  	0.50	    & \textbf{0.30} &  0.32	   & 0.32 & 1.17   & 1.00      &  \textbf{0.82}   & \textbf{0.82}    &   8.17        &  8.05   &  4.63  	  &  \textbf{3.37}  \\
			\midrule 
			Avg.      &  3.08	        &    \textbf{3.03}       &  4.08    & 4.44 & \textbf{7.64}   & 7.83     & 8.85   & 8.61    &  16.60        & 14.28   & 11.75         &  \textbf{9.28}   \\
			\bottomrule   		   
		\end{tabular}
	\end{subtable}
	
	\vspace*{0.2cm}
	\begin{subtable}[t]{0.99\textwidth}
		\centering
		\caption{Unknown-sensor protocol.}
		\label{tab:cross_sensor}
		\begin{tabular}{c | c c c c |c c c c |c c c c} 
			\toprule
			& \multicolumn{4}{c|}{\textbf{FV encoding}} & \multicolumn{4}{c|}{\textbf{Vlad encoding}} & \multicolumn{4}{c}{\textbf{BoW encoding}}  \\
			\textbf{Train set - Test set}  &  \textbf{256} & \textbf{512} & \textbf{1024} & \textbf{2048}  &  \textbf{256} & \textbf{512} & \textbf{1024} & \textbf{2048} &  \textbf{256} & \textbf{512} & \textbf{1024} & \textbf{2048}\\
			\midrule
			Bio11 - Ital11       	 &      \textbf{18.00}    &  19.10       &    19.90	    & 18.50 &  29.90		 &  23.80    &     20.10         &    \textbf{19.30}             &      \textbf{31.90}    &  32.30       &    32.92      &    38.80     \\
			Bio13 - Ital13      		 &       1.80    &   1.70       &     1.50      & \textbf{1.20}  &   \textbf{3.10}	     &   3.60   &      4.00          &     3.50             &      36.95    &  \textbf{29.65}       &    32.20      &    37.95       \\
			Ital11 - Bio11      		 &      11.50    & \textbf{9.60} &     9.90      &    10.50       &  20.00        &  17.70    &     16.00         &    \textbf{15.20}             &      32.30    &  36.75       &    \textbf{32.25}      &    39.25     \\
			Ital13 - Bio13      		 &       1.10    & \textbf{0.90} &     1.10 		&     1.10       &   1.60        &  \textbf{ 1.40}   &      1.60          &     1.90   			 &      12.30    &      9.30    &     6.75      &     \textbf{5.70}     \\ 
			\midrule
			Avg.                                 &       8.10	 & 7.83 &     8.10	    & \textbf{7.83}  &  12.90	     &  11.63   &	10.43         &	    \textbf{9.98}        &      28.36	 &  27.00	    &    \textbf{26.04}      &	 30.43      \\
			\bottomrule
		\end{tabular}
	\end{subtable}
	
	\vspace*{0.2cm}
	\begin{subtable}[t]{0.99\textwidth}
		\centering
		\caption{Cross-database protocol.}
		\label{tab:cross_database}
		\begin{tabular}{c | c c c c |c c c c |c c c c} 
			\toprule
			& \multicolumn{4}{c|}{\textbf{FV encoding}} & \multicolumn{4}{c|}{\textbf{Vlad encoding}} & \multicolumn{4}{c}{\textbf{BoW encoding}} \\
			\textbf{Train set - Test set}  &  \textbf{256} & \textbf{512} & \textbf{1024} & \textbf{2048}  &  \textbf{256} & \textbf{512} & \textbf{1024} & \textbf{2048} &  \textbf{256} & \textbf{512} & \textbf{1024} & \textbf{2048}\\
			\midrule
			Bio11 - Bio13   &     11.60     &   11.10		    &     7.00      &  \textbf{6.80} &     26.60         &      29.10        &         18.20  &     \textbf{15.70}     &   43.8 5        &    38.55      &    \textbf{28.80}       &     32.45      \\
			Bio13 - Bio11   &     12.50	  &  \textbf{11.20} &    11.60      &   12.70        & 	   \textbf{10.20}	     &     10.50        &         10.60  &     11.10     &   \textbf{49.50}         &    49.90      &   49.80        &     49.85    \\
			Ital11 - Ital13       &      9.80     &    7.60         &     6.20      &  \textbf{5.60} &     10.00	     &     \textbf{8.40} 	     &         10.10  &      8.70     &   48.75         &    48.20      &   47.75        &     \textbf{45.60}      \\
			Ital13 - Ital11       &     26.00     &   19.40		    &    13.30      &  \textbf{11.50}&     26.30 		 &     27.70        &         22.40  &     \textbf{18.10}     &   49.90         &    49.75      &   \textbf{49.60}        &     49.75     \\ 
			\midrule
			Avg.                          &     14.98     &   12.33         &     9.53      & \textbf{9.15}  &     18.28	     & 
			18.93		 &	15.33         &	    \textbf{13.40}	  &   48.00	    	&    46.60      &    \textbf{43.99}	     &   44.41     \\
			\bottomrule
		\end{tabular}
	\end{subtable}
\end{table*}

\begin{table}[t]
	\centering
	\caption{Computational performance in seconds on LivDet 2015.}
	\begin{tabular}{l c c c c } 
		\toprule
		$K$                 & \textbf{256} & \textbf{512} & \textbf{1024} & \textbf{2048} \\
		\midrule
		BoW encoding                &    0.37         &      0.38             &       0.38               &        0.39        \\
		Vlad encoding                &    1.24          &      1.33          &       1.58               &        1.98      \\
		FV encoding                    &    1.17         &       1.48          &         2.11               &        3.39        \\
		\bottomrule
	\end{tabular}
	\label{tab:efficiency}\vspace*{-0.5cm}
\end{table}

In the second experiment, the performance is evaluated over the change of data collection over the same sensor (i.e., train and test over the same sensor, but acquired for LivDet 2011 and LivDet 2013, respectively). We refer to this protocol as cross-database scenario, and Table~\ref{tab:cross_database} shows the impact of $K$ on each proposed approach. As it can be observed, again the FV encoding is able to outperform both the other encoding approaches presented in this study and the top state-of-the-art results. In particular, in three out of four cases, the best peformance is achieved for $K$ = 2048. Only fo Biometrika13 - Biometrika11 the best performance is reached for $K$ = 512. 

Under these last two scenarios, the range of variability of BoW's performance is comparable to FV and Vlad. However, the ACER is multiplied by up to 4.8 times, thus making this encoding not as suitable for PAD purposes as the other two. 

In general, we have seen how different values of $K$ can impact the performance of the PAD method, and how, depending on the scenario considered, different values yield the best performance. However, an average value of $K$ = 1024 always achieved either the best performance for FV and Vlad or it is close to it. Therefore, we can conclude that, if no data is available to carefully analyse the best option, 1024 can be chosen as a sub-optimal value for $K$.

\subsection{\textbf{Computational efficiency}}

In this last set of experiments, we study the computational efficiency of the proposed image encodings for different parameter configurations. For this purpose, we select the LivDet 2015 database, since it contains the largest images. Table~\ref{tab:efficiency} shows the average performance of the proposal over different vocabulary sizes $K$. As it could be expected, different $K$ values have an impact on the average computational efficiency of the proposed methods, since the feature vector sizes depend directly on $K$. More specifically, these efficiency results indicate that higher vocabulary sizes $K$ worsen the computational efficiency of the PAD methods in many cases. On the other hand, in some cases, larger $K$ values also lead to a better detection performance. 

It should be noted that, in all cases, the efficiency values reported by BoW encoding for each parameter combination are always below 400 milliseconds, while for FV encoding they are above 1100 milliseconds. Therefore, being FV the most accurate approach, it will be interesting to improve its computation efficiency in future work in order to attain a better trade-off between detection accuracy and computational efficiency.


\small
\bibliographystyle{IEEEtran}
\bibliography{refs}

\newpage
\vfill
\begin{IEEEbiographynophoto}
{L\'azaro J. Gonz\'alez-Soler}
received his MSc degree in Mathematics and Computer Science from the University of Havana, in 2014. Afterwards, he joined Advanced Technologies Application Center (CENATAV) from Havana, Cuba to Computer Science graduate training. He is currently a PhD. student of da/sec group at the Center for Research in Security and Privacy (CRISP), Germany. His principal research interests are focused in areas of the Biometric and Machine Learning. Specifically, biometric Presentation Attack Detection for fingerprint, iris and facial characteristics.
\end{IEEEbiographynophoto}
\vfill
\begin{IEEEbiographynophoto}
{Marta Gomez-Barrero}
received her MSc degrees in Computer Science and Mathematics, and her PhD degree in Electrical Engineering, from Universidad Autonoma de Madrid, in 2011 and 2016, respectively. Since 2016 she is a PostDoctoral researcher at the Center for Research in Security and Privacy (CRISP), Germany. Her current research focuses on the development of privacy-enhancing biometric technologies as well as Presentation Attack Detection methods, within the wider fields of pattern recognition and machine learning. She has been actively involved in international projects dealing with vulnerability evaluation of biometric systems, including the EU FP7 projects Tabula Rasa and BEAT, or the BATL project within the US IARPA Odin Program. She is also the recipient of a number of distinctions, including: EAB European Biometric Industry Award 2015, Best Ph.D. Thesis Award by Universidad Autonoma de Madrid 2015/16, Siew-Sngiem Best Paper Award at ICB 2015, Archimedes Award for young researches from Spanish Ministry of Education in 2013 and Best Poster Award at ICB 2013.
\end{IEEEbiographynophoto}
\vfill
\begin{IEEEbiographynophoto}
{Leonardo Chang}
received his bachelor degree with honors from CUJAE University in Havana, Cuba in 2007, and his M.Sc. and Ph.D. in Computer Science from the National Institute for Astrophysics, Optics, and Electronics of Mexico in 2010 and 2015, respectively. He was a Researcher at CENATAV, Cuba during 2007-2017. Currently, he is a full-time Researcher and Professor at Tecnologico de Monterrey, Mexico. His research interests include biometrics, object recognition, and video-surveillance applications. He has published several papers in top journals and conferences.
\end{IEEEbiographynophoto}
\vfill
\begin{IEEEbiographynophoto}
{Airel P\'erez-Su\'arez}
is an Associate Researcher in the Data Mining Department at the Advanced Technologies Application Centre (CENATAV), Cuba. He received his BSc in Computer Science from the Havana University in 2002. In July 2008 he obtained the MSc degree in Computational Sciences at the National Institute of Astrophysics, Optics and Electronics (INAOE) and, in July 2011, he received his PhD degree in Computational Sciences from the same institution. He is member of the Cuban Society of Mathematics and Computation and of the Cuban Association for Pattern Recognition since 2005. Dr. Airel P\'erez Su\'arez has focused his researches on clustering, but his interests also cover other areas of Data Mining such as Supervised Classification, Frequent Patterns, Association Rules, Emerging Patterns and Community detection on Social Networks.
\end{IEEEbiographynophoto}
\vfill
\begin{IEEEbiographynophoto}
{Christoph Busch}
received the Diploma degree from the Technical University of Darmstadt (TUD), Darmstadt, Germany, and the Ph.D. degree in computer graphics from TUD, in 1997. He joined the Fraunhofer Institute for Computer Graphics, Darmstadt, in 1997. He is a member of the Faculty of Computer Science and Media Technology with the Norwegian University of Science and Technology, Norway, and holds a joint appointment with the Faculty of Computer Science, Hochschule Darmstadt. Furthermore, he lectures a course on biometric systems with DTU in Copenhagen since 2007. His research includes pattern recognition, multimodal and mobile biometrics, and privacy enhancing technologies for biometric systems. He is Cofounder of the European Association for Biometrics and convener of WG3 in ISO/IEC JTC1 SC37 on Biometrics. He coauthored over 400 technical papers, and has been a speaker at international conferences.
\end{IEEEbiographynophoto}
\enlargethispage{-5in}




\end{document}